\documentclass[10pt,twocolumn,letterpaper]{article}

\usepackage[pagenumbers]{cvpr} 

\usepackage[dvipsnames]{xcolor}
\usepackage{times}
\usepackage{amsmath}
\usepackage{amssymb}
\usepackage{arydshln}
\usepackage{bm}
\usepackage{boldline}
\usepackage{booktabs}
\usepackage{color}
\usepackage{enumitem}
\usepackage{float}
\usepackage{graphicx}
\usepackage{multirow}
\usepackage{nicefrac}

\newcommand{\method}{OMS\xspace}

\usepackage{amsmath,amsfonts,bm}

\def\eqref#1{equation~\ref{#1}}

\def\1{\bm{1}}

\def\eps{{\epsilon}}

\def\rvepsilon{{\mathbf{\epsilon}}}

\def\rvv{{\mathbf{v}}}

\def\rvx{{\mathbf{x}}}

\def\rvz{{\mathbf{z}}}

\DeclareMathAlphabet{\mathsfit}{\encodingdefault}{\sfdefault}{m}{sl}
\SetMathAlphabet{\mathsfit}{bold}{\encodingdefault}{\sfdefault}{bx}{n}

\newcommand{\KL}{D_{\mathrm{KL}}}

\definecolor{cvprblue}{rgb}{0.21,0.49,0.74}
\usepackage[pagebackref,breaklinks,colorlinks,citecolor=cvprblue]{hyperref}
\usepackage{adjustbox}
\usepackage[ruled,vlined]{algorithm2e}
\SetKwInput{KwInput}{Input}               
\SetKwInput{KwOutput}{Output}
\SetKwInput{KwRequire}{Require}
\usepackage{amsmath}
\newtheorem{theorem}{Theorem}[section]
\newtheorem{lemma}[theorem]{Lemma}
\newtheorem{property}[theorem]{Property}

\usepackage{multirow}
\usepackage{subcaption}

\def\paperID{5123} 
\def\confName{CVPR}
\def\confYear{2024}

\title{One More Step: A Versatile Plug-and-Play Module for Rectifying Diffusion Schedule Flaws and Enhancing Low-Frequency Controls}

\author{
    Minghui Hu$^\dagger$\; Jianbin Zheng$^\star$\; Chuanxia Zheng$^\ddagger$\; Chaoyue Wang$^\mathsection$\; Dacheng Tao$^\mathsection$\; Tat-Jen Cham$^\dagger$ \\
    $^\dagger$Nanyang Technological University,\; $^\ddagger$University of Oxford\; \\$^\star$South China University of Technology,\; $^\mathsection$The University of Sydney  \\
    \texttt{\small \{e200008, astjcham\}@ntu.edu.sg, jabir.zheng@outlook.com, cxzheng@robots.ox.ac.uk}
}

\begin{document}

\twocolumn[
{
\maketitle
\vspace*{-1.5cm}
\begin{figure}[H]
    \centering
    \hsize=\textwidth
    \includegraphics[width=\textwidth]{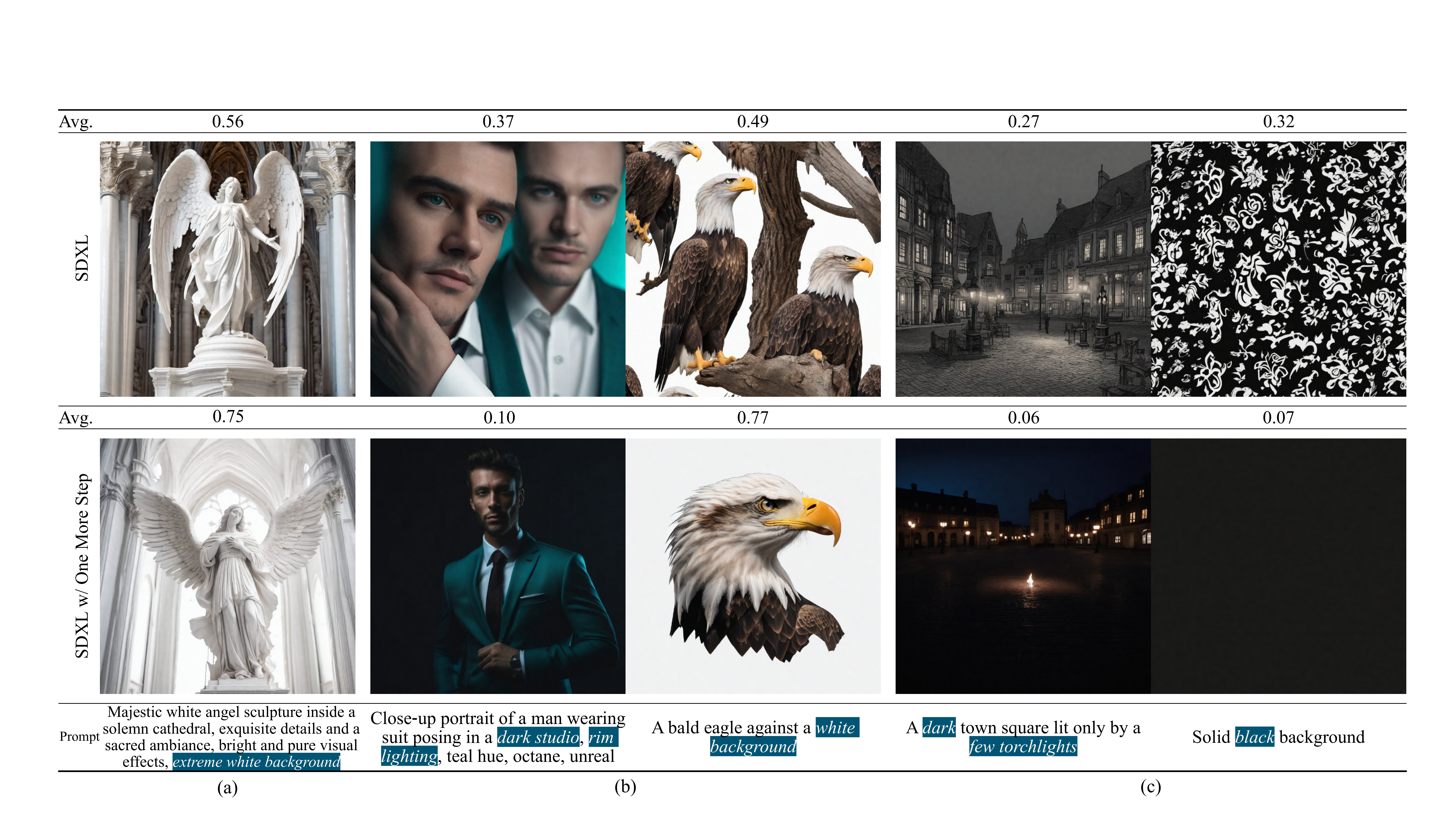}
    \vspace{-1.06cm}
    \caption{\textbf{Example results of our One More Step method on various sceceries.}
    Traditional sampling methods (Top row) not only lead to (a) generated images converging towards the mean value, but also cause (b) the structure of generated objects to be chaotic, or (c) the theme to not follow prompts. 
    Our proposed \textit{One More Step} addresses these problems effectively \emph{without modifying any parameters} in the pre-trained models.
    \textbf{Avg.} denotes the average pixel value of the images, which are normalized to fall within the range of [0, 1].
    }
    \label{fig:front}
\end{figure}
}
]
\maketitle

\begin{abstract}
\vspace{-0.4cm}

It is well known that many open-released foundational diffusion models have difficulty in generating images that substantially depart from average brightness, despite such images being present in the training data.
This is due to an inconsistency: while denoising starts from pure Gaussian noise during inference, the training noise schedule retains residual data even in the final timestep distribution, due to difficulties in numerical conditioning in mainstream formulation, leading to unintended bias during inference.
To mitigate this issue, certain $\epsilon$-prediction models are combined with an ad-hoc offset-noise methodology. 
In parallel, some contemporary models have adopted zero-terminal SNR noise schedules together with $\mathbf{v}$-prediction, 
which necessitate major alterations to pre-trained models. 
However, such changes risk destabilizing a large multitude of community-driven applications anchored on these pre-trained models. 
In light of this, our investigation revisits the fundamental causes, leading to our proposal of an innovative and principled remedy, called One More Step (OMS). 
By integrating a compact network and incorporating an additional simple yet effective step during inference, OMS elevates image fidelity and harmonizes the dichotomy between training and inference, while preserving original model parameters. Once trained, various pre-trained diffusion models with the same latent domain can share the same OMS module. Codes and models are released at \href{https://jabir-zheng.github.io/OneMoreStep/}{here}.
\end{abstract}    
\section{Introduction}
\label{sec:intro}

Diffusion models have emerged as a foundational method for improving quality, diversity, and resolution of generated images~\cite{ho2020denoising, sohl2015deep}, due to the robust generalizability and straightforward training process.
At present, a series of open-source diffusion models, exemplified by Stable Diffusion~\cite{rombach2022high}, hold significant sway and are frequently cited within the community. 
Leveraging these open-source models, numerous researchers and artists have either directly adapted~\cite{zhang2023adding, hu2023cocktail} or employed other techniques~\cite{hu2021lora} to fine-tune and craft an array of personalized models. 

However, recent findings by~\citet{lin2023common, karras2022elucidating} identified deficiencies in existing noise schedules, leading to generated images primarily characterized by medium brightness levels.
Even when prompts include explicit color orientations, the generated images tend to gravitate towards a mean brightness. 
Even when prompts specify “a solid black image” or “a pure white background”, the models will still produce images that are obviously incongruous with the provided descriptions (see examples in~\cref{fig:front}). 
We deduced that such inconsistencies are caused by a divergence between inference and training stages, 
due to inadequacies inherent in the dominant noise schedules. 
In detail, during the inference procedure, the initial noise is drawn from a \emph{pure Gaussian distribution}. 
In contrast, during the training phase, previous approaches such as linear~\cite{ho2020denoising} and cosine~\cite{nichol2021improved} schedules manifest a non-zero SNR at the concluding timestep. 
This results in low-frequency components, especially the mean value, of the training dataset remaining residually present in the final latents during training, to which the model learns to adapt.
However, when presented with pure Gaussian noise during inference, the model behaves as if these residual components are still present, resulting in the synthesis of suboptimal imagery~\cite{chen2023importance, hoogeboom2023simple}.

In addressing the aforementioned issue,~\citet{guttenberg2023diffusion} first proposed a straightforward solution: 
introducing a specific offset to the noise derived from sampling, thereby altering its mean value. 
This technique has been designated as \textit{offset noise}. 
While this methodology has been employed in some of the more advanced models~\cite{podell2023sdxl}, 
it is \emph{not} devoid of inherent challenges. 
Specifically, the incorporation of this offset disrupts the iid distribution characteristics of the noise across individual units. 
Consequently, although this modification enables the model to produce images with high luminance or profound darkness, it might inadvertently generate signals incongruent with the distribution of the training dataset. 
A more detailed study~\cite{lin2023common} suggests a \textit{zero terminal SNR} method that rescaling the model's schedule to ensure the SNR is zero at the terminal timestep can address this issue. 
Nonetheless, this strategy necessitates the integration of $\mathbf{v}$-prediction models~\cite{salimans2022progressive} and mandates subsequent fine-tuning across the entire network, regardless of whether the network is based on $\mathbf{v}$-prediction or $\epsilon$-prediction~\cite{ho2020denoising}. 
Besides, fine-tuning these widely-used pre-trained models would render many community models based on earlier releases incompatible, diminishing the overall cost-to-benefit ratio.

To better address this challenge, we revisited the reasons for its emergence: 
\emph{flaws in the schedule result in a mismatch between the marginal distributions of terminal noise during the training and inference stages}.
Concurrently, we found the distinct nature of this terminal timestep: the latents predicted by the model at the terminal timestep continue to be associated with the data distribution.

Based on the above findings, we propose a plug-and-play method, named \textbf{O}ne \textbf{M}ore \textbf{S}tep, that solves this problem without necessitating alterations to the pre-existing trained models, as shown in~\cref{fig:front}. 
This is achieved by training an auxiliary text-conditional network tailored to
map pure Gaussian noise to the data-adulterated noise assumed by the pre-trained model, optionally under the guidance of an additional prompt, and is introduced prior to the inception of the iterative sampling process. 

OMS can rectify the disparities in marginal distributions encountered during the training and inference phases. 
Additionally, it can also be leveraged to adjust the generated images through an additional prompt, due to its unique property and position in the sampling sequence. 
It is worth noting that our method exhibits versatility, being amenable to any variance-preserving~\cite{song2020score} diffusion framework, irrespective of the network prediction type, whether $\epsilon$-prediction or $\mathbf{v}$-prediction, and independent of the SDE or ODE solver employed. Experiments demonstrate that SD1.5, SD2.1, LCM~\cite{luo2023latent} and other popular community models can \emph{share the same} OMS module for improved image generation.

\section{Preliminaries}
\label{sec:preliminaries}

\subsection{Diffusion Model and its Prediction Types}
We consider diffusion models~\cite{sohl2015deep,ho2020denoising} specified in discrete time space and variance-preserving (VP)~\cite{song2020score} formulation. 
Given the training data $\rvx \in p(\rvx)$, a diffusion model performs the forward process to destroy the data $\rvx_0$ into noise $\rvx_T$ according to the pre-defined variance schedule $\{\mathbf{\beta}_t\}_{t=1}^{T}$ according to a perturbation kernel, defined as:

\begin{equation}
    q(\rvx_{1:T}| \rvx_0) := \prod_{t=1}^T q(\rvx_t | \rvx_{t-1}),
\end{equation}
\begin{equation}
    q(\rvx_t| \rvx_{t-1}) := \mathcal{N}\left(\rvx_t;\sqrt{1-\beta_t}\rvx_{t-1} , \beta_t \mathbf{I}\right).
\end{equation}
The forward process also has a closed-form equation, which allows directly sampling $x_t$ at any timestep $t$ from $x_0$:

\begin{equation}
    q(\rvx_t|\rvx_0) :=  \mathcal{N} (\rvx_t; \sqrt{\bar{\alpha}_t}\rvx_0, (1-\bar{\alpha}_t)\mathbf{I}),
    \label{sample_x_t}
\end{equation}
where $\bar{\alpha}_t = \prod_{s=1}^t \alpha_s$ and $\alpha_t = 1-\beta_t$. Furthermore, the \textit{signal-to-noise ratio} (SNR) of the latent variable can be defined as:

\begin{equation}
    \text{SNR}(t) = \bar{\alpha}_t / (1-\bar{\alpha}_t).
\end{equation}
The reverse process denoises a sample $\rvx_T$ from a standard Gaussian distribution to a data sample $\rvx_0$ following:

\begin{equation}
    p_{\theta} (\rvx_{t-1} | \rvx_t) := \mathcal{N} (\rvx_{t-1}; \tilde{\mu}_t, \tilde{\sigma}_t^2 \mathbf{I}).
\label{p_sample}
\end{equation}
\begin{equation}
    \tilde{\mu}_t := \frac{\sqrt{\bar{\alpha}_{t-1}}\beta_t}{1-\bar{\alpha}_t} \rvx_0 + \frac{\sqrt{\alpha_t}(1-\bar{\alpha}_{t-1})}{1-\bar{\alpha}_t} \rvx_t
\label{mu}
\end{equation}
Instead of directly predicting $\tilde{\mu}_t$ using a network $\theta$, predicting the reparameterised $\epsilon$ for $\rvx_0$ leads to a more stable result~\cite{ho2020denoising}:

\begin{equation}
    \tilde{\rvx}_0 :=  (\rvx_t - \sqrt{1-\bar{\alpha}_t}\epsilon_{\theta}(\rvx_t, t) ) / \sqrt{\bar{\alpha}_t}
\label{eps_mu}
\end{equation}
and the variance of the reverse process $\tilde{\sigma}_t^2$ is set to be $\sigma_t^2 = \frac{1-\bar{\alpha}_{t-1}}{1-\bar{\alpha}_t}\beta_t$ while $\rvx_t \sim \mathcal{N}(0,1)$. 
Additionally, predicting velocity~\cite{salimans2022progressive} is another parameterisation choice for the network to predict:

\begin{equation}
    \rvv_t := \sqrt{\bar{\alpha}_t} \epsilon -  \sqrt{1-\bar{\alpha}_t}\rvx_0;
\label{v_rep}
\end{equation}
which can reparameterise $\tilde{\rvx}_0$ as:

\begin{equation}
    \tilde{\rvx}_0 :=  \sqrt{\bar{\alpha}_t} \rvx_t - \sqrt{1-\bar{\alpha}_t}\rvv_{\theta}(\rvx_t, t) 
    \label{v_mu}
\end{equation}

\subsection{Offset Noise and Zero Terminal SNR}

\textit{Offset noise}~\cite{guttenberg2023diffusion} is a straightforward method to generate dark or light images more effectively by fine-tuning the model with modified noise. 
Instead of directly sampling a noise from standard Gaussian Distribution $\epsilon \sim \mathcal{N}(0, \mathbf{I})$, one can sample the initial noise from

\begin{equation}
    \epsilon \sim \mathcal{N} (0, \mathbf{I} + 0.1 \boldsymbol{\Sigma}),
\end{equation}
where $\boldsymbol{\Sigma}$ is a covariance matrix of all ones, representing fully correlated dimensions. 
This implies that the noise bias introduced to pixel values across various channels remains consistent. 
In the initial configuration, the noise attributed to each pixel is independent, devoid of coherence. By adding a common noise across the entire image (or along channels), changes can be coordinated throughout the image, facilitating enhanced regulation of low-frequency elements. 
However, this is an unprincipled ad hoc adjustment that inadvertently leads to the noise mean of inputs deviating from representing the mean of the actual image.

A different research endeavor proposes a more fundamental approach to mitigate this challenge~\cite{lin2023common}: 
rescaling the beta schedule ensures that the low-frequency information within the sampled latent space during training is thoroughly destroyed. 
To elaborate, current beta schedules are crafted with an intent to minimize the SNR at $\rvx_T$. 
However, constraints related to model intricacies and numerical stability preclude this value from reaching zero.
Given a beta schedule used in LDM~\cite{rombach2022high}:

\begin{equation}
    \beta_t = \left( \sqrt{0.00085} \frac{T-t}{T-1} + \sqrt{0.012} \frac{t-1}{T-1} \right) ^2,
\label{sd_schedule}
\end{equation}
the terminal SNR at timestep $T=1000$ is 0.004682 and $\sqrt{\bar{\alpha}_T}$ is 0.068265.
To force terminal SNR=0, rescaling can be done to make $\bar{\alpha}_T = 0$ while keeping $\bar{\alpha}_0$ fixed. 
Subsequently, this rescaled beta schedule can be used to fine-tune the model to avoid the information leakage. 
Concurrently, to circumvent the numerical instability induced by the prevalent $\epsilon$-prediction at zero terminal SNR, this work mandates the substitution of prediction types across all timesteps with $\mathbf{v}$-prediction.
However, such approaches cannot be correctly applied for sampling from pre-trained models that are based on Eq.~\ref{sd_schedule}.
\section{Methods}

\subsection{Discrepancy between Training and Sampling}

From the beta schedule in Eq.~\ref{sd_schedule}, we find the SNR  \emph{cannot} reach zero at terminal timestep as $\bar{\alpha}_T$ is not zero. Substituting the value of $\bar{\alpha}_T$ in Eq.~\ref{sample_x_t}, we can observe more intuitively that during the training process, the latents sampled by the model at $T$ deviate significantly from expected values:

\begin{equation}
    \rvx_T^{\mathcal{T}} = \sqrt{\bar{\alpha}_T^{\mathcal{T}} } \rvx_0 + \sqrt{1-\bar{\alpha}_T^{\mathcal{T}} } \rvz,
\end{equation}
where $\sqrt{\bar{\alpha}_T^{\mathcal{T}} } = 0.068265$ and $\sqrt{1-\bar{\alpha}_T^{\mathcal{T}} } = 0.997667$.

During the training phase, the data fed into the model is not entirely pure noise at timestep $T$. 
It contains minimal yet data-relevant signals. 
These inadvertently introduced signals contain low-frequency details, such as the overall mean of each channel. 
The model is subsequently trained to denoise by respecting the mean in the leaked signals. 
However, in the inference phase, \emph{sampling is executed using standard Gaussian distribution}.
Due to such an inconsistency in the distribution between training and inference, when given the zero mean of Gaussian noise, the model unsurprisingly produces samples with the mean value presented at $T$, resulting in the manifestation of images with median values. 
Mathematically, the directly sampled variable $\rvx_T^{\mathcal{S}}$ in the inference stage adheres to the standard Gaussian distribution $\mathcal{N}(0, \mathbf{I})$. 
However, the marginal distribution of the forward process from image space $\mathcal{X}$ to the latent space $\rvx_T^{\mathcal{T}}$ during training introduces deviations of the low-frequency information, which is non-standard Gaussian distribution.

\begin{figure}[ht!]
    \centering
    \includegraphics[width=0.8\linewidth]{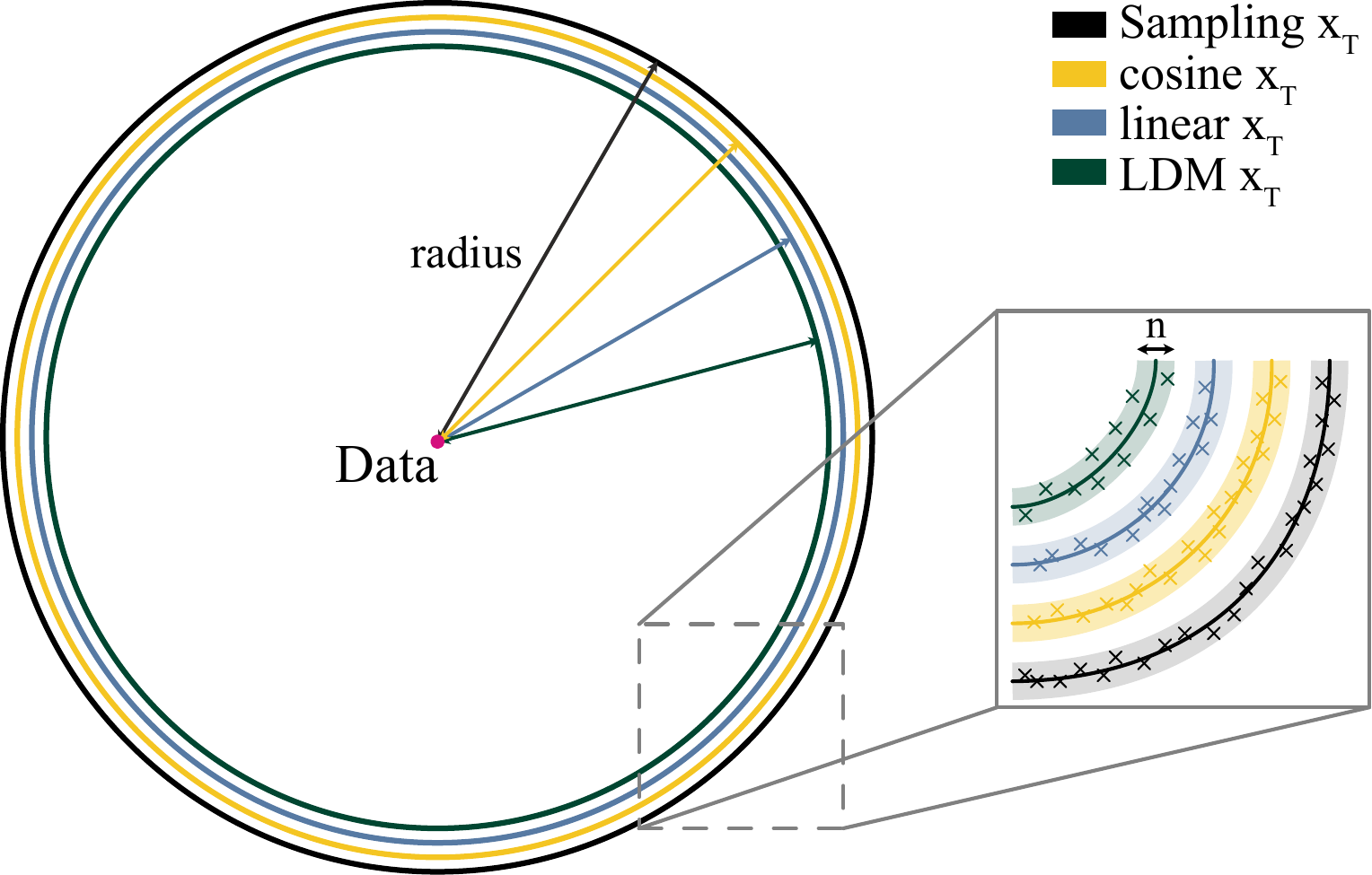}
    \caption{The geometric illustration of concentration mass in the equatorial cross-section of high-dimensional Gaussians, where its mass concentrates in a very small annular band around the radius. Different colors represent the results sampled based on different schedules. It can be seen that as the SNR increases, the distribution tends to be more data-centric, thus the radius of the distribution is gradually decreasing.}
    \label{fig:geovis}
\end{figure}

This discrepancy is more intuitive in the visualization of high-dimensional Gaussian space by estimating the radius $r$~\cite{zhu2023boundary}, which is closely related to the expected distance of a random point from the origin of this space. 
Theoretically, given a point $\rvx = (x_1, x_2, \dots, x_d)$ sampled within the Gaussian domain spanning a $d$-dimensional space, the squared length or the norm of $\rvx$ inherently denotes the squared distance from this point to the origin according to:

\begin{equation}
    E(x_1^2 + x_2^2 + \dots + x_d^2 ) = dE(x_1^2) = d\sigma^2,
\end{equation}
and the square root of the norm is Gaussian radius $r$. 
When this distribution is anchored at the origin with its variance represented by $\sigma$, its radius in Gaussian space is determined by:

\begin{equation}
    r = \sigma \sqrt{d},
\end{equation}
the average squared distance of any point randomly selected from the Gaussian distribution to the origin.
Subsequently, we evaluated the radius within the high-dimensional space for both the variables present during the training phase $r^{\mathcal{T}}$ and those during the inference phase $r^{\mathcal{S}}$, considering various beta schedules, the results are demonstrated in~\cref{tab:radius}. 
Additionally, drawing from \cite{zhu2023boundary, blum2020foundations}, we can observe that the concentration mass of the Gaussian sphere resides above the equator having a radius magnitude of $\mathcal{O}\left(\frac{r}{\sqrt{d}}\right)$, also within an annulus of constant width and radius $n$. 
Therefore, we can roughly visualize the distribution of terminal variables during both the training and inference processes in~\cref{fig:geovis}. 
It can be observed that a discernible offset emerges between the terminal distribution $\rvx_T^{\mathcal{T}}$ and $\rvx_T^{\mathcal{S}}$ and $r^{\mathcal{S}} > r^{\mathcal{T}}$. 
This intuitively displays the discrepancy between training and inference, which is our primary objective to mitigate. 
Additional theoretical validations are relegated to the Appendix~\ref{appe_hdg} for reference.

\begin{table}[ht!]

    \centering
    \begin{adjustbox}{width=\linewidth}
    \begin{tabular}{lcccc}
    \toprule
          Schedule & SNR($T$) & $r^{\mathcal{T}}$ & $r^{\mathcal{S}}$ & $\Delta r$\\
    \midrule
    cosine  &2.428e-09   &443.404205     &443.404235 & 3.0518e-05\\
    linear  &4.036e-05   &443.393676     &443.399688 & 6.0119e-03\\
    LDM Pixels    &4.682e-03   &442.713593     &443.402527 & 6.8893e-01\\
    LDM Latents$^\dag$     &4.682e-03   &127.962364     &127.996811 & 3.4447e-02\\
    \bottomrule
    \end{tabular}
    \end{adjustbox}
    \footnotesize{$^\dag$ LDMs were conducted both in the \textit{unit variance} latent space (4*64*64) and pixel space (3*256*256) while others are conducted in pixel space.}\\
    \caption{Estimation of the Gaussian radius during the sampling and inference phases under different beta schedules.
    Here, we randomly sampled 20,000 points to calculate the radius.
    }
    \label{tab:radius}
\end{table}

\subsection{Prediction at Terminal Timestep}
\label{ptt}

According to Eq.~\ref{p_sample}~\&~\ref{eps_mu}, we can obtain the sampling process under the text-conditional DDPM pipeline with $\eps$-prediction at timestep $T$:

\begin{equation}
    \rvx_{T-1} = \frac{1}{\sqrt{\alpha_T}} \left( \rvx_T - \frac{1-\alpha_T}{\sqrt{1-\bar{\alpha}_T}} \eps_{\theta}  \right) + \sigma_T \rvz,
\end{equation}
where $\rvz, \rvx_T \sim \mathcal{N}(0,\mathrm{I})$. In this particular scenario, it is obvious that the ideal SNR($T$) = 0 setting (with $\alpha_T$ = 0) will lead to numerical issues, and any predictions made by the network at time $T$ with an SNR($T$) = 0 are arduous and lack meaningful interpretation. 
This also elucidates the necessity for the linear schedule to define its start and end values~\cite{ho2020denoising} and for the cosine schedule to incorporate an offset $s$~\cite{nichol2021improved}.

Utilizing SNR-independent $\mathbf{v}$-prediction can address this issue. By substituting Eq.~\ref{v_mu} into Eq.~\ref{p_sample}, we can derive:

\begin{equation}
    \rvx_{T-1} = \sqrt{\alpha_T} \rvx_T - \frac{\sqrt{\bar{\alpha}_{T-1}}(1-\alpha_T)}{\sqrt{1-\bar{\alpha}_T}}\rvv_{\theta} + \sigma_T \rvz,
\end{equation}
which the assumption of SNR($T$) = 0 can be satisfied: when SNR($T$) = 0, the reverse process of calculating $\rvx_{T-1}$ depends only on the prediction of $\rvv_{\theta}(\rvx_T,T)$,

\begin{equation}
    \rvx_{T-1} = -\sqrt{\bar{\alpha}_{T-1}}\rvv_{\theta} + \sigma_T \rvz,
    \label{sde_s}
\end{equation}
which can essentially be interpreted as predicting the direction of $\rvx_0$ according to Eq.~\ref{v_rep}: 

\begin{equation}
    \rvx_{T-1} = \sqrt{\bar{\alpha}_{T-1}}\rvx_0 + \sigma_T \rvz.
\end{equation}
This is also consistent with the conclusions of angular parameterisation\footnote{Details about $\mathbf{v}$-prediction and angular parametersation can be found in the Appendix.~\ref{appe_ddim_phi}.}~\cite{salimans2022progressive}. To conclude, under the ideal condition of SNR = 0, the model is essentially forecasting the L2 mean of the data, hence the objective of the $\mathbf{v}$-prediction at this stage aligns closely with that of the direct $\rvx_0$-prediction. Furthermore, this prediction by the network at this step is independent of the pipeline schedule, implying that the prediction remain consistent irrespective of the variations in noise input.

\subsection{Adding One More Step}

Holding the assumption that $\rvx_T$ belongs to a standard Gaussian distribution, the model actually has no parameters to be trained with pre-defined beta schedule, so the objective $L_T$ should be the constant:

\begin{equation}
    L_T = \KL \left(q(\rvx_T|\rvx_0) \Vert p(\rvx_T)\right).
\end{equation}
In the present architecture, the model conditioned on $\rvx_T$ actually does not participate in the training. However, existing models have been trained to predict based on $\rvx_T^{\mathcal{T}}$, 
which indeed carries some data information. 

\begin{figure}[t!]
    \centering
    \includegraphics[width=\linewidth]{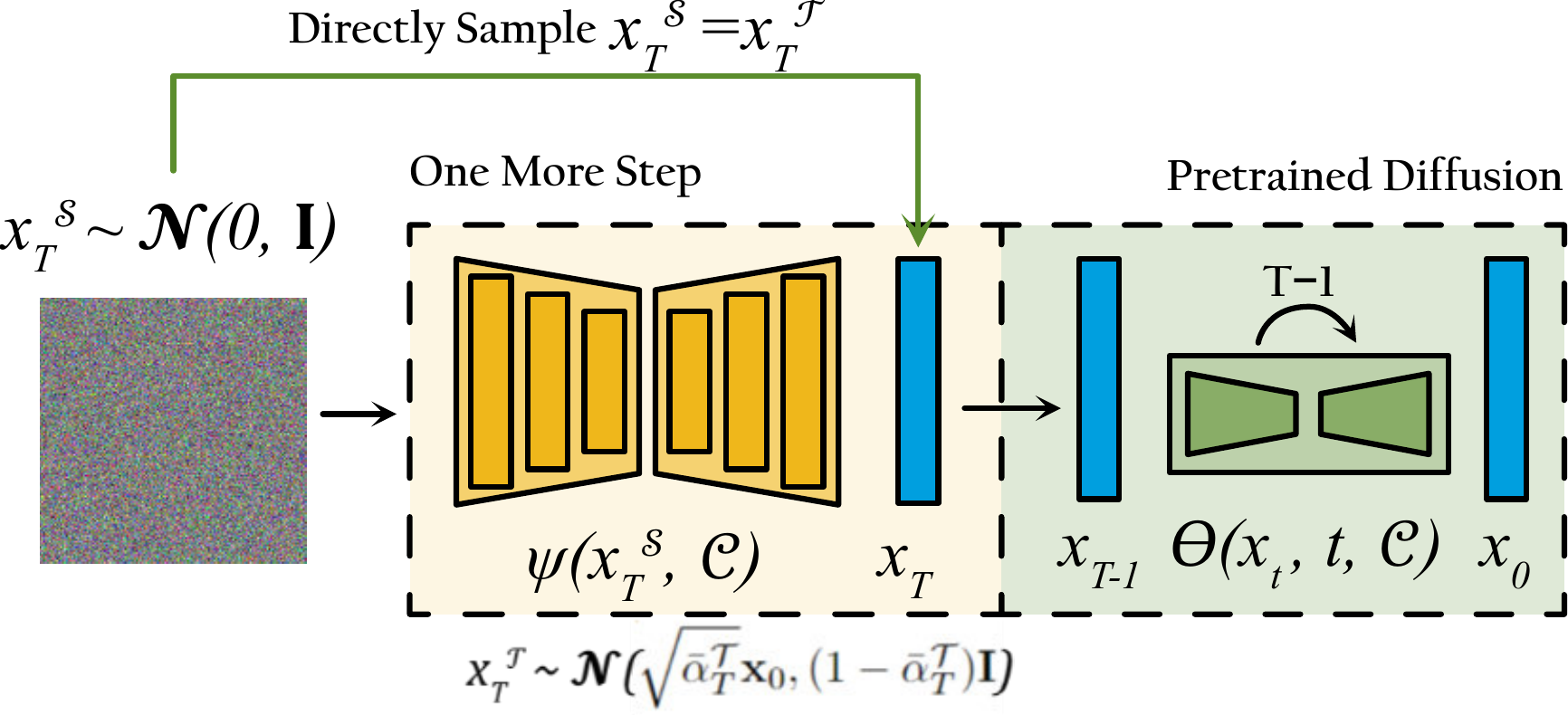}
    \caption{The pipeline of \textit{\textbf{O}ne \textbf{M}ore \textbf{S}tep}. The section highlighted in {\color{Dandelion}{\textbf{yellow}}} signifies our introduced OMS module, with $\psi$ being the only trainable component. The segments in {\color{RoyalBlue}{\textbf{blue}}} represents latent vectors, and {\color{ForestGreen}{\textbf{green}}} represents the pre-trained model used only for the inference.}
    \label{fig:omspp}
\end{figure}

Drawing upon prior discussions, we know that the model's prediction conditioned on $\rvx_T^{\mathcal{S}}$ should be the average of the data, which is also independent of the beta schedule. 
This understanding brings a new perspective to the problem: retaining the whole pipeline of the current model, encompassing both its parameters and the beta schedule.
In contrast, we can reverse $\rvx_T^{\mathcal{S}}$ to  $\rvx_T^{\mathcal{T}}$ by introducing \textbf{O}ne \textbf{M}ore \textbf{S}tep (\textbf{OMS}). 
In this step, we first train a network $\psi(\rvx_T^{\mathcal{S}}, \mathcal{C})$ to perform $\mathbf{v}$-prediction conditioned on $\rvx_T^{\mathcal{S}} \sim \mathcal{N}(0, \mathbf{I})$ with L2 loss $\Vert \rvv_T^\mathcal{S} - \tilde\rvv_T^{\mathcal{S}} \Vert^2_2$, where $\rvv_T^\mathcal{S} = -\rvx_0$ and $\tilde\rvv_T^\mathcal{S}$ is the prediction from the model.
Next, we reconstruct $\tilde{\rvx}_T^{\mathcal{T}}$ based on the output of $\psi$ with different solvers. 
In addition to the SDE Solver delineated in Eq.~\ref{sde_s}, we can also leverage prevalent ODE Solvers, \eg, DDIM~\cite{song2020denoising}:

\begin{equation}
    \tilde{\rvx}_{T}^{\mathcal{T}} =  \sqrt{\bar{\alpha}_T^{\mathcal{T}}} \tilde{\rvx}_0 + \sqrt{1-\bar{\alpha}_T^{\mathcal{T}} - \sigma_T^2} \rvx_T^{\mathcal{S}} + \sigma_T \rvz,
    \label{omsddim}
\end{equation}
where $\tilde{\rvx}_0$ is obtained based on $\psi(\rvx_T^{\mathcal{S}}, \mathcal{C})$.
Subsequently, $\tilde{\rvx}_{T}^{\mathcal{T}}$ can be utilized as the initial noise and incorporated into various pre-trained models. From a geometrical viewpoint, we employ a model conditioned on $\rvx_T^{\mathcal{S}}$ to predict $ \tilde{\rvx}_{T}^{\mathcal{T}}$ that aligns more closely with $\mathcal{N} \left( \sqrt{\bar{\alpha}_T^{\mathcal{T}}}\rvx_0, (1-\bar{\alpha}_T^{\mathcal{T}}) \mathbf{I}\right)$, which has a smaller radius and inherits to the training phase of the pre-trained model at timestep $T$. The whole pipeline and geometric explanation is demonstrated in Figs.~\ref{fig:omspp}~\&~\ref{fig:omsgeo}, and the detailed algorithm and derivation can be referred to Alg.~\ref{alg1} in Appendix~\ref{sec:detailed_alg}. 

Notably, the prompt $\mathcal{C}_{\psi}$ in OMS phase $\psi(\cdot)$ can be different from the conditional information $\mathcal{C}_{\theta}$ for the pre-trained diffusion model $\theta(\cdot)$. 
Modifying the prompt in OMS phase allows for additional manipulation of low-frequency aspects of the generated image, such as color and luminance. 
Besides, OMS module also support classifier free guidance~\cite{ho2022classifier} to strength the text condition:

\begin{equation}
    \psi_\text{cfg} ({\rvx_T^{\mathcal{S}}, \mathcal{C}_{\psi}, \emptyset, \omega_\psi}) = \psi ({\rvx_T^{\mathcal{S}}, \emptyset}) + \omega_\psi \left( \psi  ({\rvx_T^{\mathcal{S}}, \mathcal{C}}_{\psi}) - \psi ({\rvx_T^{\mathcal{S}}, \emptyset}) \right),  
\end{equation}
where $\omega_\psi$ is the CFG weights for OMS. Experimental results for inconsistent prompt and OMS CFG can be found in Sec.~\ref{sec_incos}. 

\begin{figure}[ht!]
    \centering
    \includegraphics[width=0.5\linewidth]{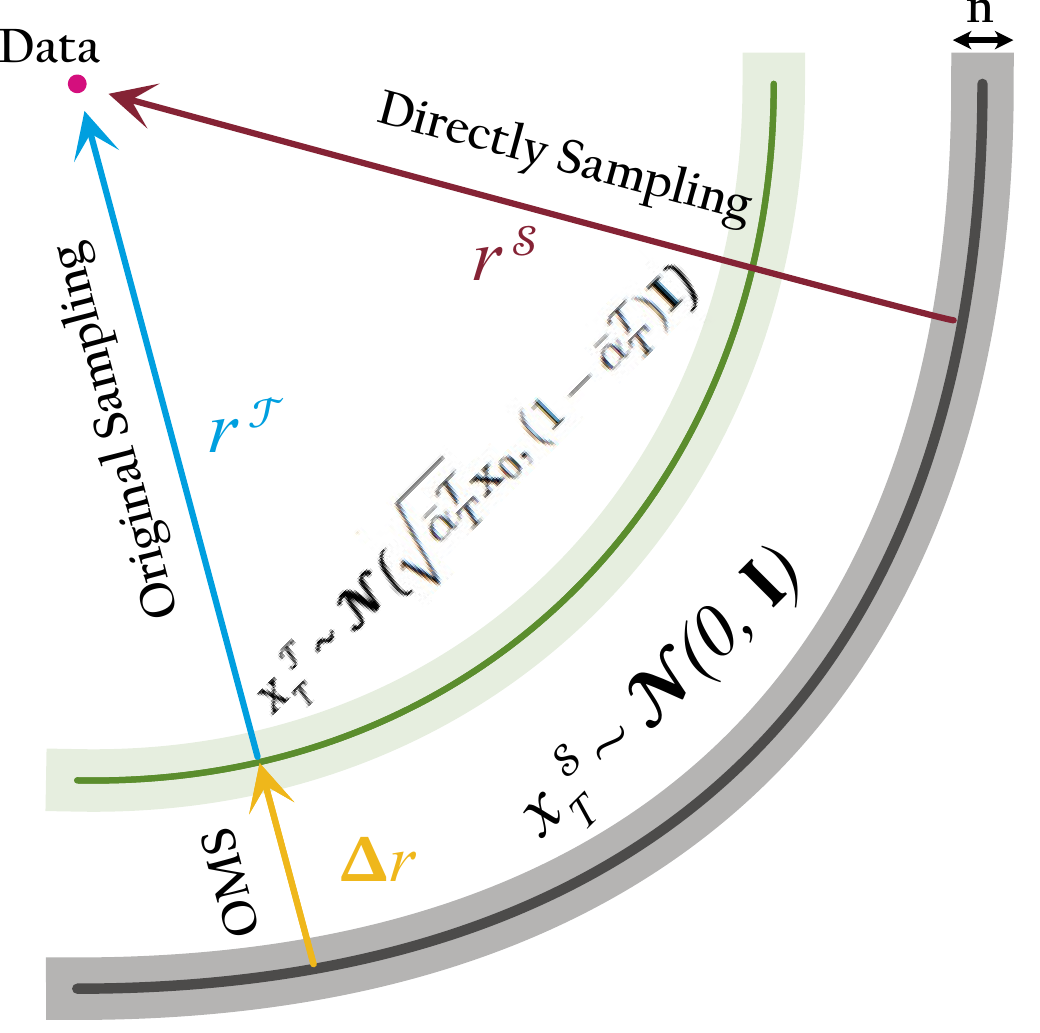}
    \caption{Geometric explanation of \textit{\textbf{O}ne \textbf{M}ore \textbf{S}tep}. While directly sampling method requires sampling from a Gaussian distribution with a radius of $r^{\mathcal{T}}$, yet it samples from the standard Gaussian with $r^{\mathcal{S}}$ in practice. OMS bridges the gap $\Delta r$ between $r^{\mathcal{S}}$ and the required $r^{\mathcal{T}}$ through an additional inference step. Here $n$ is the width of the narrow band where the distribution mass is concentrated.}
    \label{fig:omsgeo}
\end{figure}

It is worth noting that OMS can be adapted to any pre-trained model within the same space.  
Simply put, our OMS module trained in the same VAE latent domain can adapt to any other model that has been trained within the same latent space and data distribution. 
Details of the OMS and its versatility can be found in Appendix~\ref{adco}~\&~\ref{vae_con}.

\section{Experiments}

\begin{figure*}
    \centering
        \begin{subfigure}[b]{\linewidth}
        \includegraphics[width=\linewidth]{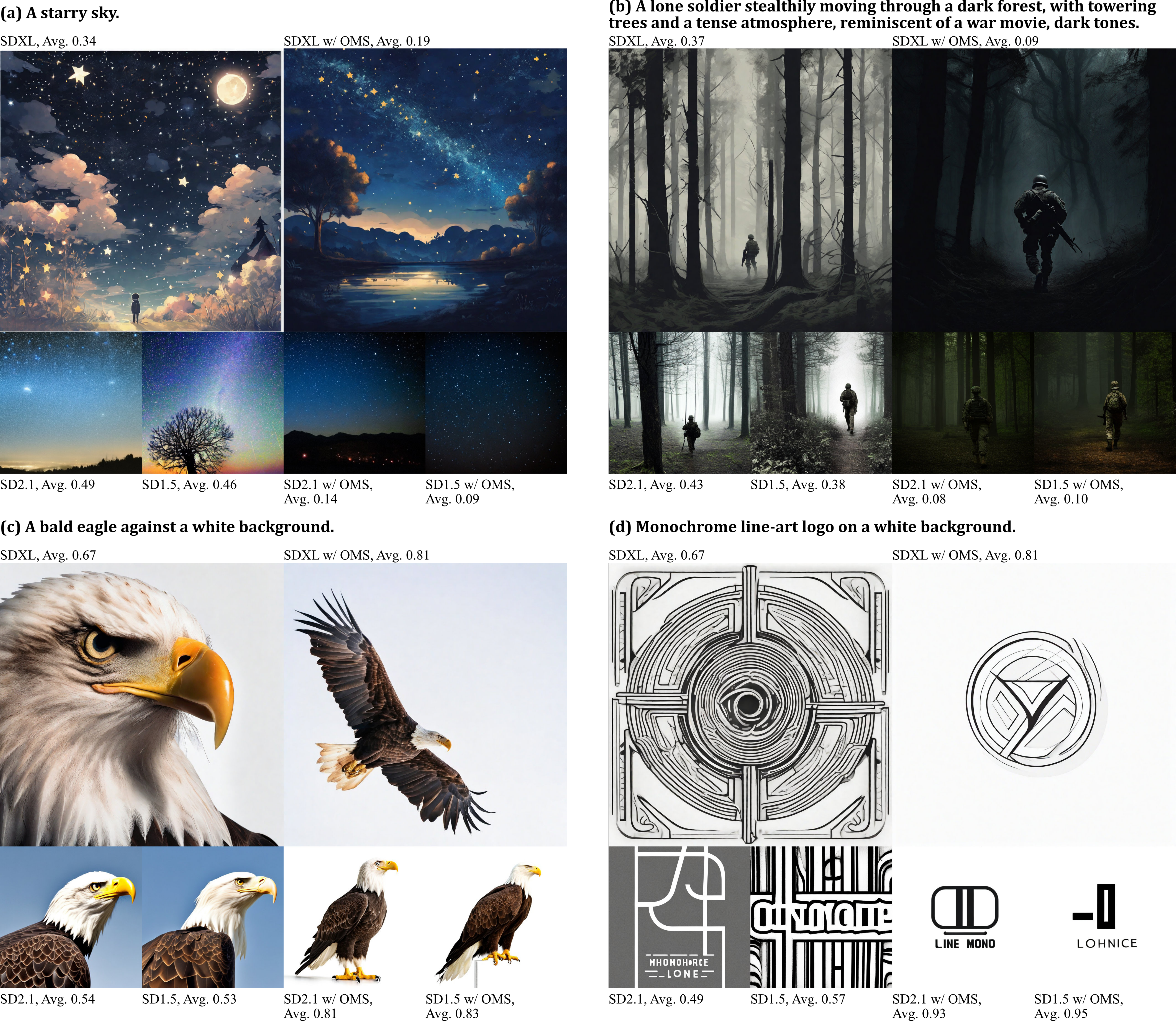}
        \caption{Images are sampled by DDIM with $50$+$1$ Steps.}
        \end{subfigure}

        \begin{subfigure}[b]{\linewidth}
        \includegraphics[width=\linewidth]{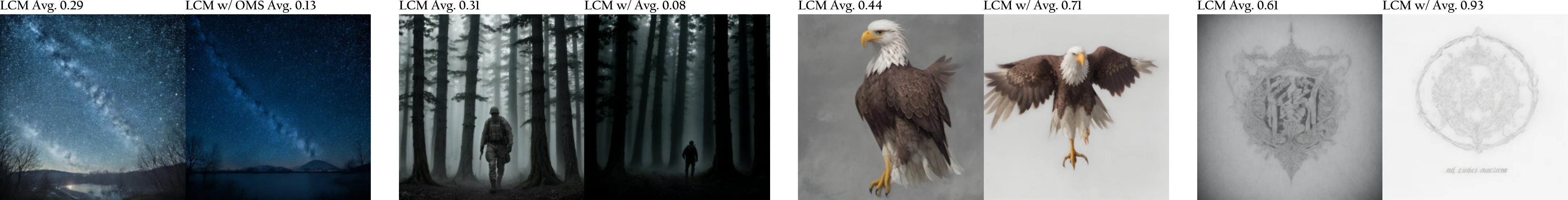}
        \caption{Images are sampled by LCM with $4$+$1$ Steps and the same prompts sets.}
        \end{subfigure}
    
    \caption{Qualitative comparison. 
    For each image pair, the left shows results from original pre-trained diffusion models, whereas the right demonstrates the output from these same models enhanced with the \method under identical prompts. 
    It is worth noting that SD1.5, SD2.1~\cite{rombach2022high} and LCM~\cite{luo2023latent}
    in this experiment~\emph{share the same OMS module, rather than training an exclusive module for each one.
    }.
    }
    \label{fig:quality}
\end{figure*}

This section begins with an evaluation of the enhancements provided by the proposed \method module to pre-trained generative models, 
examining both qualitative and quantitative aspects, and its adaptability to a variety of diffusion models.
Subsequently, we conducted ablation studies on pivotal designs and dive into several interesting occurrences.

\subsection{Implementation Details}
We trained our \method module on LAION 2B dataset~\cite{schuhmann2022laion}. 
OMS module architecture follows the widely used UNet~\cite{ronneberger2015u, ho2020denoising} in diffusion, and we evaluated different configurations, e.g., number of layers. 
By default we employ OpenCLIP ViT-H to encode text for the OMS module and trained the model for 2,000 steps. For detailed implementation information, please refer to the Appendix.~\ref{appe_imple}.

\subsection{Performance}

\paragraph{Qualitative}

\cref{fig:front,fig:quality} illustrate that our approach is capable of producing images across a large spectrum of brightness levels. 
Among these, SD1.5, SD2.1 and LCM~\cite{luo2023latent} use the \textit{same OMS module}, whereas SDXL employs a separately trained OMS module\footnote{The VAE latent domain of the SDXL model differs considerably from those of SD1.5, SD2.1 and LCM. For more detailed information, please refer to the Appendix.~\ref{vae_con}}. 
As shown in the~\cref{fig:quality} left, existing models invariably yield samples of medium brightness and are \textit{not} able to generate accurate images when provided with explicit prompts. 
In contrast, our model generates a distribution of images that is more broadly covered based on the prompts. 
In addition to further qualifying the result, we also show some integration of the widely popular customized LoRA \cite{hu2021lora} and base models in the community with our module in Appendix.~\ref{sec:extre}, which also ascertains the versatility of OMS. 

\begin{figure}
    \centering
    \includegraphics[width=\linewidth]{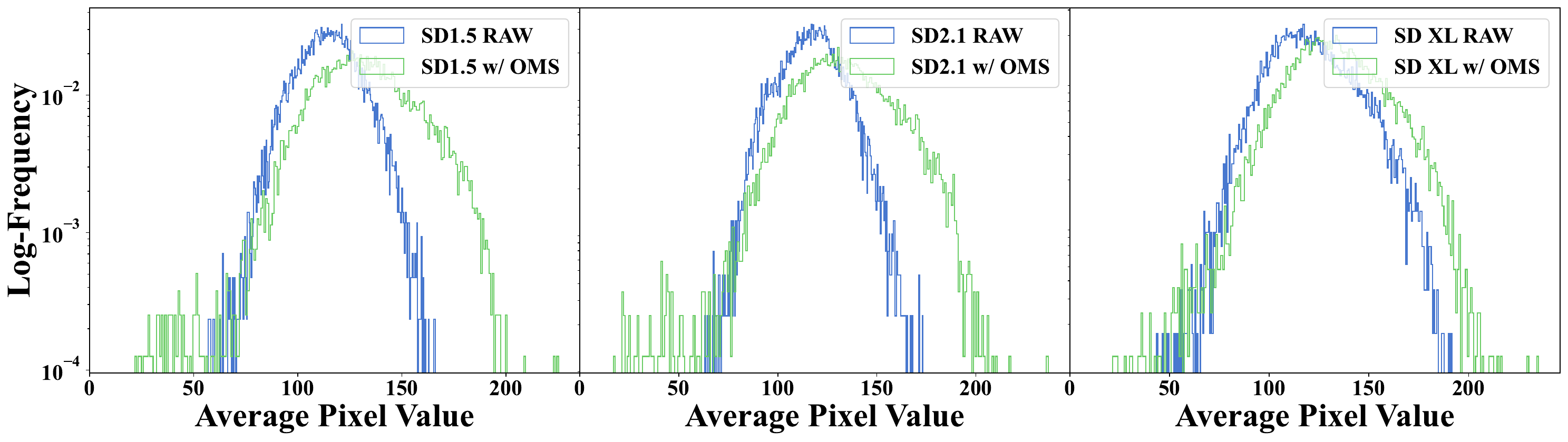}
    \caption{Log-frequency histogram of image mean values.}
    \label{fig:hist}
\end{figure}

\paragraph{Quantitative}
For the quantitative evaluation, we randomly selected 10k captions from MS COCO~\cite{lin2014microsoft} for zero-shot generation of images. 
We used Fr\'echet Inception Distance (FID), CLIP Score~\cite{radford2021learning}, Image Reward~\cite{xu2023imagereward}, and PickScore~\cite{kirstain2023pick} to assess the quality, text-image alignment, and human preference of generated images.
\cref{tab:fid} presents a comparison of these metrics across various models, either with or without the integration of the OMS module. 
It is worth noting that~\citet{kirstain2023pick} demonstrated that the FID score for COCO zero-shot image generation has a \emph{negative correlation} with visual aesthetics, thus the FID metric is not congruent with the goals of our study.
Instead, we have further computed the Precision-Recall (PR)~\cite{kynkaanniemi2019improved} and Density-Coverage (DC)~\cite{naeem2020reliable} between the ground truth images and those generated, as detailed in the~\cref{tab:fid}.
Additionally, we calculate the mean of images and the Wasserstein distance~\cite{rubner2000earth}, and visualize the log-frequency distribution in \cref{fig:hist}.
It is evident that our proposed OMS module promotes a more broadly covered distribution.

\begin{table*}
\centering
\resizebox{\linewidth}{!}{%
\begin{tabular}{lcccccccccc}
\toprule
Model                                      &     & FID $\downarrow$ & CLIP Score$\uparrow$ & ImageReward$\uparrow$ & PickScore$\uparrow$ & Precision$\uparrow$ & Recall$\uparrow$ & Density$\uparrow$ & Coverage$\uparrow$ & Wasserstein$\downarrow$  \\ \midrule
\multicolumn{1}{l}{\multirow{2}{*}{SD1.5}} & RAW & \textbf{12.52}    & 0.2641 & 0.1991 & 21.49 & 0.60 & \textbf{0.55} & 0.56 & 0.54 & 22.47 \\
\multicolumn{1}{l}{}                       & OMS & 14.74    & \textbf{0.2645} & \textbf{0.2289} & \textbf{21.55} & \textbf{0.64} & 0.46 & \textbf{0.64} & \textbf{0.57} & \textbf{7.84} \\\midrule
\multirow{2}{*}{SD2.1}                     & RAW & \textbf{14.10}    & 0.2624 & 0.4501 & 21.80 & 0.58 & \textbf{0.55} & 0.52 & 0.50 & 21.63 \\
                                           & OMS & 15.72    & \textbf{0.2628} & \textbf{0.4565} & \textbf{21.82} & \textbf{0.61} & 0.48 & \textbf{0.58} & \textbf{0.54} & \textbf{7.70} \\\midrule
\multirow{2}{*}{SD XL}                     & RAW & \textbf{13.14}    & 0.2669 & 0.8246 & 22.51 & 0.64 & \textbf{0.52} & 0.67 & 0.63 & 11.08 \\
                                           & OMS & 13.29    & \textbf{0.2679} & \textbf{0.8730} & \textbf{22.52} & \textbf{0.65} & 0.49 & \textbf{0.70} & \textbf{0.64} & \textbf{7.25} \\ \bottomrule
\end{tabular}
}
\caption{Quantitative evaluation. All models use DDIM sampler with 50 steps, guidance weight $\omega_\theta=7.5$ and negative prompts are $\emptyset$. For OMS module, there is no OMS CFG $\omega_\psi=1$ and no inconsistent prompt $\mathcal{C}_\psi = \mathcal{C}_\theta$. Better results are highlighted in \textbf{bold}. }
\label{tab:fid}
\end{table*}

\begin{figure}[ht!]
    \centering
    \begin{subfigure}[b]{0.9\linewidth}
            \caption{Modifying the prompts in the OMS module can adjust the brightness in the generated images.}
        \label{fig:light}
        \includegraphics[width=\linewidth]{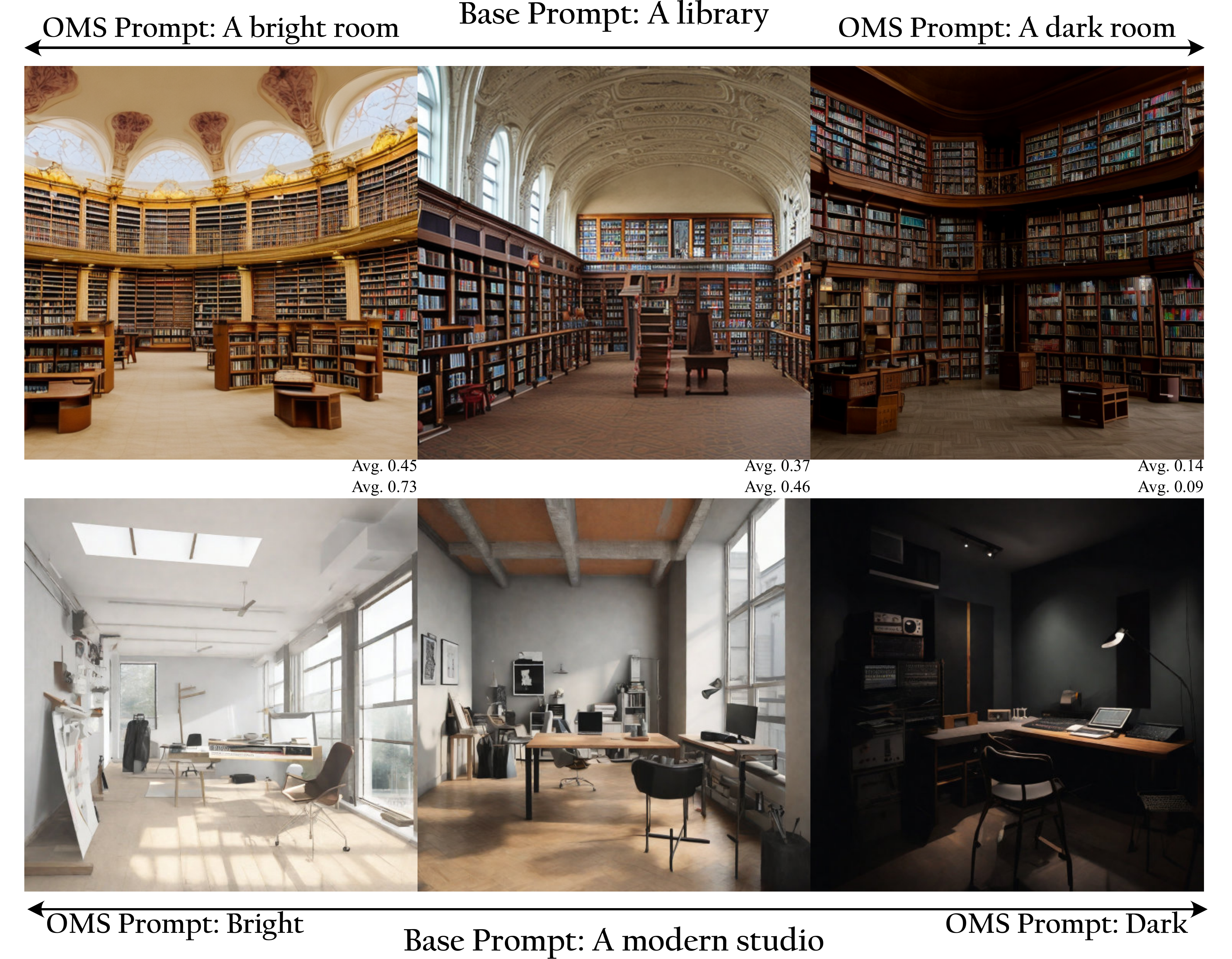}

    \end{subfigure}
    \begin{subfigure}[b]{0.9\linewidth}
            \caption{Modifying the prompts in the OMS module can change the object color in the generated images.}
        \label{fig:color}
        \includegraphics[width=\linewidth]{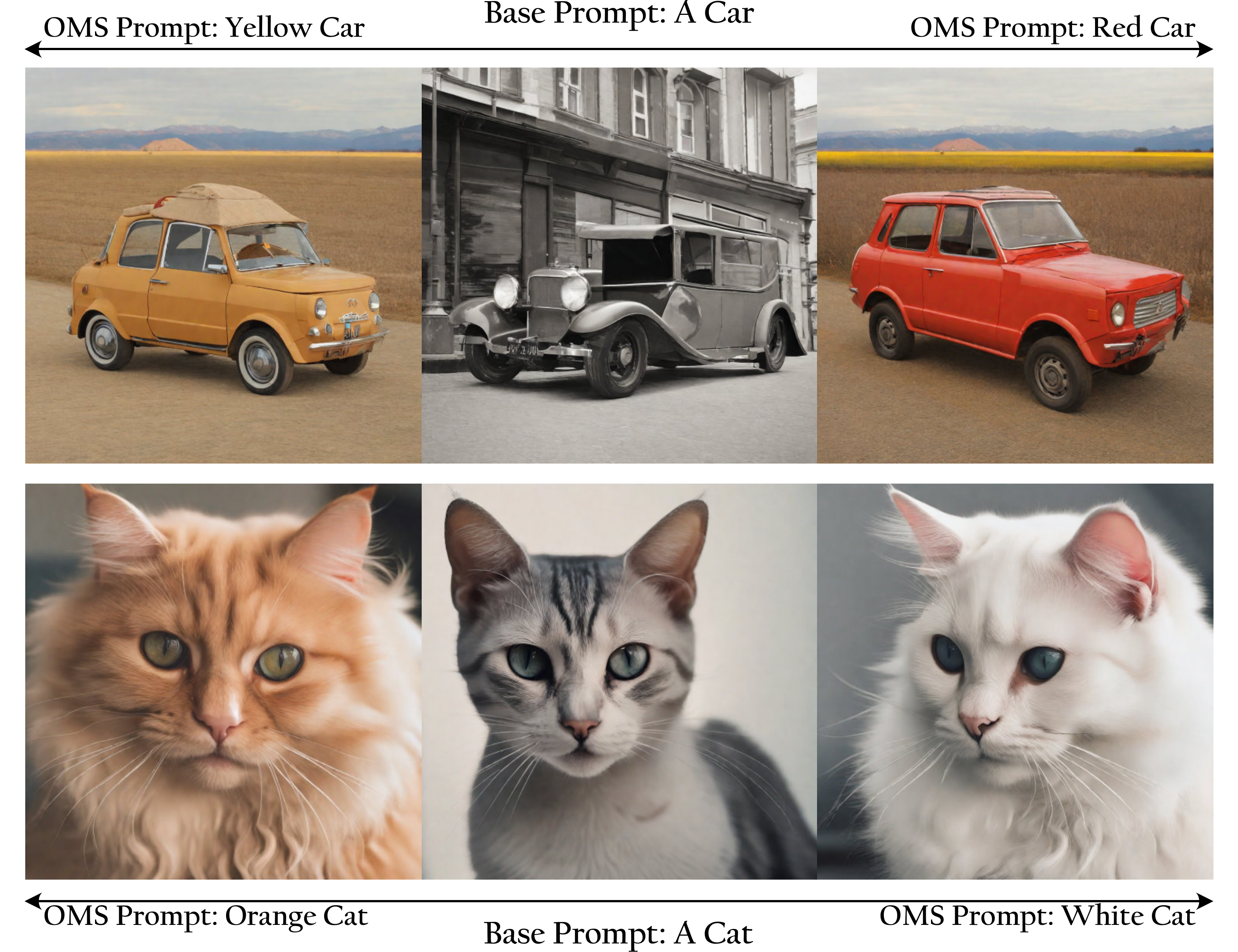}

    \end{subfigure}
    \caption{Altering the prompts in the OMS module, while keeping the text prompts in the diffusion backbone model constant, can notably affect the characteristics of the images generated.}
    \label{fig:both_figures}
\end{figure}

\begin{figure*}[ht!]
    \centering
    \includegraphics[width=0.95\textwidth]{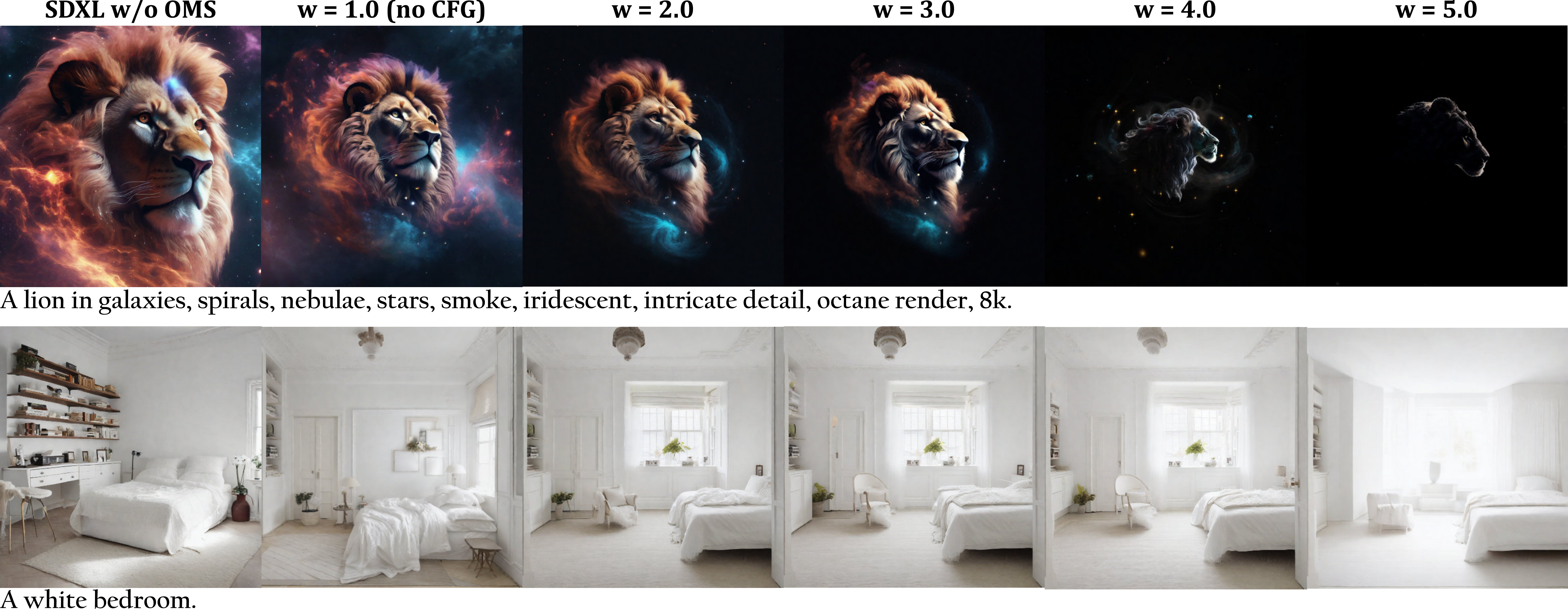}
    \caption{Images under the same prompt but with different OMS CFG weights applied in OMS module. Notably, CFG weight of the pre-trained diffusion model remains 7.5.}
    \label{fig:omscfg}
\end{figure*}

\subsection{Ablation}

\paragraph{Module Scale}
Initially, we conducted some research on the impact of model size. 
The aim is to explore whether variations in the parameter count of the OMS model would influence the enhancements in image quality. 
We experimented with OMS networks of three different sizes and discovered that the amelioration of image quality is \emph{not} sensitive to the number of OMS parameters. 
From~\cref{tab:size_text} in Appendix~\ref{appe_imple}, we found that even with only 3.7M parameters, the model was still able to successfully improve the distribution of generated images. 
This result offers us an insight: it is conceivable that during the entire denoising process, certain timesteps encounter relatively trivial challenges, hence the model scale of specific timestep might be minimal and using a Mixture of Experts strategy~\cite{balaji2022ediffi} but with different scale models at diverse timesteps may effectively reduce the time required for inference.

\vspace{-0.3cm}
\paragraph{Text Encoder}
Another critical component in OMS is the text encoder. 
Given that the OMS model’s predictions can be interpreted as the mean of the data informed by the prompt, it stands to reason that a more potent text encoder would enhance the conditional information fed into the OMS module. 
However, experiments show that the improvement brought by different encoders is also limited. 
We believe that the main reason is that OMS is only effective for low-frequency information in the generation process, and these components are unlikely to affect the explicit representation of the image. 
The diverse results can be found in~\cref{tab:size_text} in Appendix~\ref{appe_imple}.

\vspace{-0.3cm}
\paragraph{Modified Prompts}
\label{sec_incos}
In addition to providing coherent prompts, we also conducted experiments to examine the impact of the low-frequency information during the OMS step with different prompts, mathematically $\mathcal{C}_\psi \neq \mathcal{C}_\theta$. 
We discovered that the brightness level of the generated images can be easily controlled with terms like $\mathcal{C}_\psi$ is ``dark'' or ``light'' in the OMS phase, as can be seen from~\cref{fig:light}. 
Additionally, our observations indicate that the modified prompts used in the OMS are capable of influencing other semantic aspects of the generated content, including color variations as shown in~\cref{fig:color}.

\vspace{-0.3cm}
\paragraph{Classifier-free guidance}
Classifier-free guidance (CFG) is well-established for enhancing the quality of generated content and is a common practice~\cite{ho2022classifier}. CFG still can play a key component in OMS, effectively influencing the low-frequency characteristics of the image in response to the given prompts. Due to the unique nature of our OMS target for generation, the average value under $\emptyset$ is close to that of conditioned ones $\mathcal{C}_\psi$. As a result, even minor applications of CFG can lead to considerable changes. Our experiments show that a CFG weight $\omega_\psi=2$ can create distinctly visible alterations.  In~\cref{fig:omscfg}, we can observe the performance of generated images under different CFG weights for OMS module. It worth noting that CFG weights of OMS and the pre-trained model are imposed independently.
\vspace{-0.2cm}
\section{Conclusion}
\vspace{-0.1cm}
In summary, our observations indicate a discrepancy in the terminal noise between the training and sampling stages of diffusion models due to the schedules, resulting in a distribution of generated images that is centered around the mean. To address this issue, we introduced \textit{\textbf{O}ne \textbf{M}ore \textbf{S}tep}, which adjusts for the training and inference distribution discrepancy by integrating an additional module while preserving the original parameters. Furthermore, we discovered that the initial stages of the denoising process with low SNR largely determine the low-frequency traits of the images, particularly the distribution of brightness, and this phase does not demand an extensive parameter set for accurate model fitting.
{
    \small
    \bibliographystyle{ieeenat_fullname}
    \bibliography{main}
}
\appendix
\clearpage
\setcounter{page}{1}
\maketitlesupplementary

\begin{figure*}[ht!]
    \centering
        \includegraphics[width=\textwidth]{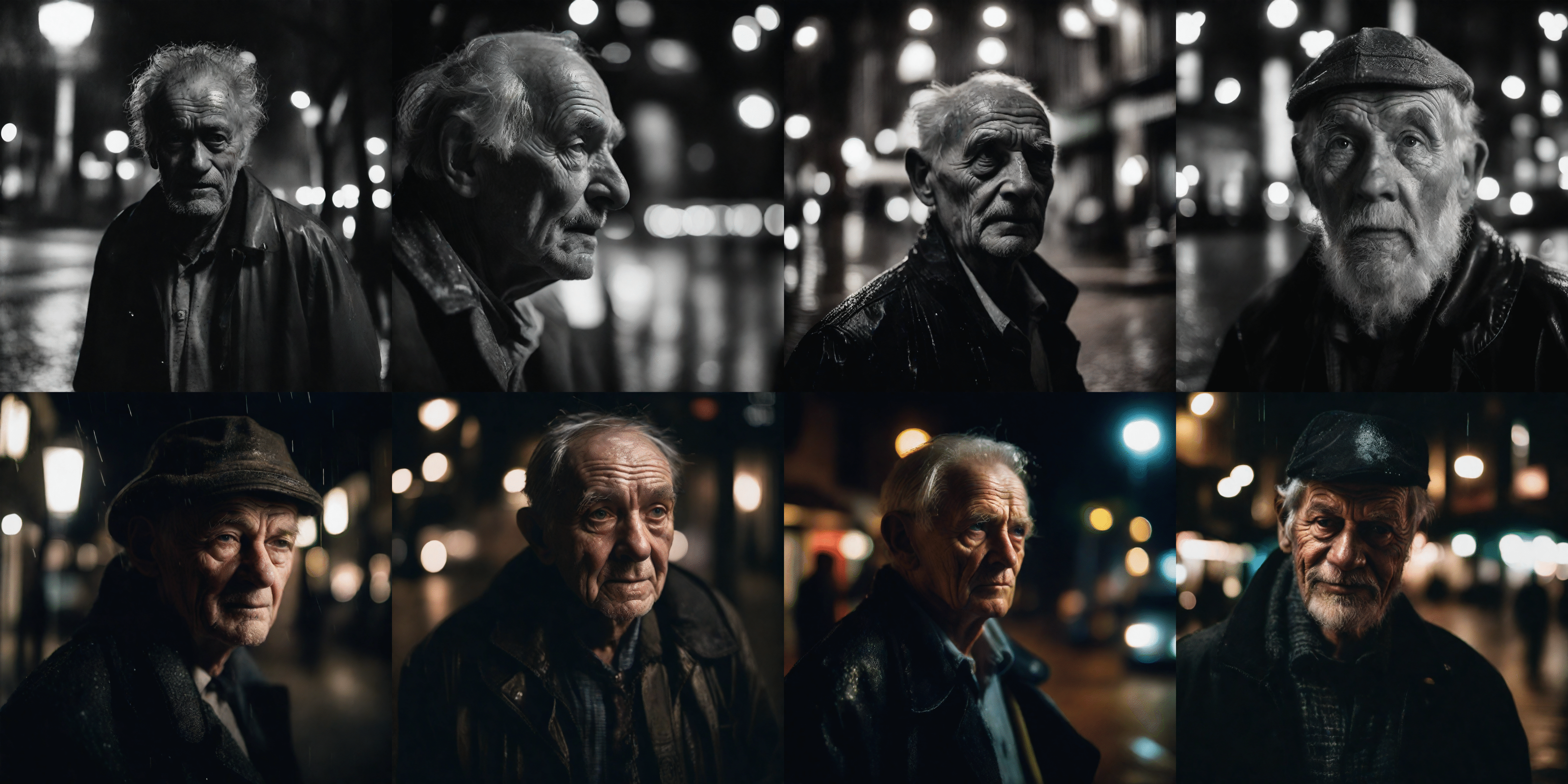}
        \caption{The same set of configurations (SDXL w/ LCM-LoRA with $4$(+$1$) Steps) as~\cref{fig:lcm_lora_rep} but with different random seeds. SDXL with LCM-LoRA leans towards black-and-white images, but OMS produces more  colorful images. It is worth noting the mean value of all SDXL with LCM-LoRA results is 0.24 while the average value of OMS results is \textbf{0.17}. We hypothesize the tendency of SDXL to produce black-and-white images is a direct result of flaws in its scheduler for training.}

\end{figure*}

\section{Related Works}

Diffusion models~\cite{ho2020denoising, song2020score} have significantly advanced the field of text-to-image synthesis~\cite{nichol2021glide, ramesh2022hierarchical, rombach2022high, balaji2022ediffi, hu2022global, hu2022unified}. These models often operate within the latent space to optimize computational efficiency~\cite{rombach2022high} or initially generate low-resolution images that are subsequently enhanced through super-resolution techniques~\cite{ramesh2022hierarchical, balaji2022ediffi}. Recent developments in fast sampling methods have notably decreased the diffusion model's generation steps from hundreds to just a few~\cite{song2020denoising, lu2022dpm, karras2022elucidating, liu2022pseudo, luo2023latent}. Moreover, incorporating classifier guidance during the sampling phase significantly improves the quality of the results~\cite{dhariwal2021diffusion}. While classifier-free guidance is commonly used~\cite{ho2022classifier}, exploring other guidance types also presents promising avenues for advancements in this domain~\cite{zhao2022egsde, graikos2022diffusion}.

\section{High Dimensional Gaussian}
\label{appe_hdg}

In our section, we delve into the geometric and probabilistic features of high-dimensional Gaussian distributions, which are not as evident in their low-dimensional counterparts. These characteristics are pivotal for the analysis of latent spaces within denoising models, given that each intermediate latent space follows a Gaussian distribution during denoising. Our statement is anchored on the seminal work by~\cite{blum2020foundations, zhu2023boundary}. These works establish a connection between the high-dimensional Gaussian distribution and the latent variables inherent in the diffusion model.

\begin{property}
        For a unit-radius sphere in high dimensions, as the dimension $d$ increases, the volume of the sphere goes to 0, and the maximum possible distance between two points stays at 2.
\label{property:radius}
\end{property}
\begin{lemma}
The surface area $A(d)$ and the volume $V(d)$ of a unit-radius sphere in $d$-dimensions can be obtained by:
\begin{equation}
    A(d) = \frac{2 \pi^{d/2}}{\Gamma(d/2)}, V(d) = \frac{\pi^{d/2}}{\frac{d}{2}\Gamma(d/2)},
\end{equation}
\label{lemma:volume}
\end{lemma}
where $\Gamma(x)$ represents an extension of the factorial function to accommodate non-integer values of $x$, the aforementioned Property~\ref{property:radius} and Lemma~\ref{lemma:volume} constitute universal geometric characteristics pertinent to spheres in higher-dimensional spaces. These principles are not only inherently relevant to the geometry of such spheres but also have significant implications for the study of high-dimensional Gaussians, particularly within the framework of diffusion models during denoising process.

\begin{property}
The volume of a high-dimensional sphere is essentially all contained in a thin slice at the equator and is simultaneously contained in a narrow annulus at the surface, with essentially no interior volume. Similarly, the surface area is essentially all at the equator.
\label{property:annulus}
\end{property}
The Property~\ref{property:annulus} implies that samples from $\rvx_T^{\mathcal{S}}$ are falling into a narrow annulus.

\begin{lemma}
For any $c>0$, the fraction of the volume of the hemisphere above the plane $x_1 = \frac{c}{\sqrt{d-1}}$ is less than $\frac{2}{c} e^{-\frac{c^2}{2}}$.
\label{lemma_equator}
\end{lemma}

\begin{lemma}
For a d-dimensional spherical Gaussian of variance 1, all but $\frac{4}{c^2}e^{-c^2/4}$ fraction of its mass is within the annulus $\sqrt{d-1}-c \leq r \leq \sqrt{d-1} + c$ for any $c > 0$.
\label{lemma_annulus}
\end{lemma}
Lemmas~\ref{lemma_equator}~\&~\ref{lemma_annulus} imply the volume range of the concentration mass above the equator is in the order of $O(\frac{r}{\sqrt{d}})$, also within an annulus of constant width and radius $\sqrt{d-1}$. Figs.\ref{fig:geovis}~\&~\ref{fig:omsgeo} in main paper illustrates the geometric properties of the ideal sampling space $\rvx_T^{\mathcal{S}}$ compared to the practical sampling spaces $\rvx_T^{\mathcal{T}}$ derived from various schedules, which should share an identical radius ideally.

\begin{property}
The maximum likelihood spherical Gaussian for a set of samples is the one over center equal to the sample mean and standard deviation equal to the standard deviation of the sample.
\label{lemma:radius}
\end{property}
The above Property~\ref{lemma:radius} provides the theoretical foundation whereby the mean of squared distances serves as a robust statistical measure for approximating the radius of high-dimensional Gaussian distributions.

\section{Expression of DDIM in angular parameterization}
\label{appe_ddim_phi}

The following covers derivation that was originally presented in \cite{salimans2022progressive}, with some corrections. We can simplify the DDIM update rule by expressing it in terms of $\phi_{t} = \text{arctan}(\sigma_{t}/\alpha_{t})$, rather than in terms of time $t$ or log-SNR $\lambda_t$, as we show here.

Given our definition of $\phi$, and assuming a variance preserving diffusion process, we have $\alpha_{\phi} = \cos(\phi)$, $\sigma_{\phi}=\sin(\phi)$, and hence $\rvz_{\phi} = \cos(\phi)\rvx + \sin(\phi)\rvepsilon$. We can now define the velocity of $\rvz_{\phi}$ as
\begin{align}
\rvv_\phi \equiv \frac{d \rvz_{\phi}}{d\phi} = \frac{d\cos(\phi)}{d\phi}\rvx + \frac{d\sin(\phi)}{d\phi}\rvepsilon =\cos(\phi)\rvepsilon - \sin(\phi)\rvx.
\end{align}
Rearranging $\rvepsilon, \rvx, \rvv$, we then get:
\begin{align}
\sin(\phi)\rvx &= \cos(\phi)\rvepsilon - \rvv_{\phi} \nonumber \\
&= \frac{\cos(\phi)}{\sin(\phi)}(\rvz - \cos(\phi)\rvx) - \rvv_{\phi}
\end{align}
\begin{align}
\sin^{2}(\phi)\rvx = \cos(\phi)\rvz - \cos^{2}(\phi)\rvx - \sin(\phi)\rvv_{\phi}
\end{align}
\begin{align}
(\sin^{2}(\phi)+\cos^{2}(\phi))\rvx = \rvx = \cos(\phi)\rvz - \sin(\phi)\rvv_{\phi},
\end{align}
and similarly we get $\epsilon = \sin(\phi) \rvz_{\phi} + \cos(\phi) \rvv_{\phi}$.

Furthermore, we define the predicted velocity as:
\begin{align}
\hat\rvv_\theta(\rvz_{\phi}) \equiv \cos(\phi)\hat\rvepsilon_\theta(\rvz_{\phi}) - \sin(\phi)\hat\rvx_\theta(\rvz_{\phi}),
\end{align}
where $\hat\rvepsilon_\theta(\rvz_{\phi}) = (\rvz_{\phi} - \cos(\phi)\hat\rvx_\theta(\rvz_{\phi}))/\sin(\phi)$.

Rewriting the DDIM update rule in the introduced terms then gives:
\begin{equation}
\begin{aligned}
\rvz_{\phi_{s}} =& \cos(\phi_{s})\hat{\rvx}_\theta(\rvz_{\phi_{t}}) + \sin(\phi_s)\hat{\rvepsilon}_{\theta}(\rvz_{\phi_{t}})\\
=&\cos(\phi_{s})(\cos(\phi_t)\rvz_{\phi_{t}} - \sin(\phi_t)\hat\rvv_\theta(\rvz_{\phi_t})) + \\
&\sin(\phi_s)(\sin(\phi_t) \rvz_{\phi_t} + \cos(\phi_t)\hat\rvv_\theta(\rvz_{\phi_t}))\\
=&[\cos(\phi_{s})\cos(\phi_t)  {\color{RoyalBlue}{\boldsymbol{+}}} \sin(\phi_s)\sin(\phi_t)]\rvz_{\phi_{t}} + \\  
&[\sin(\phi_s)\cos(\phi_t) - \cos(\phi_{s})\sin(\phi_t)]\hat\rvv_\theta(\rvz_{\phi_t}).
\end{aligned}
\end{equation}

\begin{figure}[ht]
    \centering
    \includegraphics[width=0.6\linewidth]{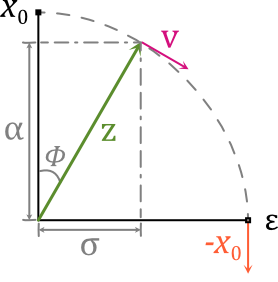}
    \caption{Visualization of reparameterizing the diffusion process in terms of $\phi$ and $\rvv_{\phi}$. We highlight the scenario where SNR is equal to zero in orange.}
    \label{fig:phi}
\end{figure}

Finally, we use the trigonometric identities
\begin{equation}
\begin{aligned}
 {\color{RoyalBlue}{\cos(\phi_{s})\cos(\phi_t) + \sin(\phi_s)\sin(\phi_t)}} &= \cos(\phi_s - \phi_t) \\ 
\sin(\phi_s)\cos(\phi_t) - \cos(\phi_{s})\sin(\phi_t) &= \sin(\phi_s - \phi_t),
\end{aligned}
\end{equation}
to find that\footnote{ The {\color{RoyalBlue}{highlighted}} part corrects minor errors that occurred in Eqs 34 and 35 from~\cite{salimans2022progressive} }
\begin{align}
\rvz_{\phi_{s}} = \cos(\phi_s - \phi_t)\rvz_{\phi_{t}} + \sin(\phi_s - \phi_t)\hat\rvv_\theta(\rvz_{\phi_t}).
\end{align}
or equivalently
\begin{align}
\rvz_{\phi_{t}-\delta} = \cos(\delta)\rvz_{\phi_{t}} - \sin(\delta)\hat\rvv_\theta(\rvz_{\phi_t}).
\end{align}
Viewed from this perspective, DDIM thus evolves $\rvz_{\phi_s}$ by moving it on a circle in the $(\rvz_{\phi_t}, \hat\rvv_{\phi_t})$ basis, along the $-\hat\rvv_{\phi_t}$ direction. 
When SNR is set to zero, the $v$-prediction effectively reduces to the $\rvx_0$-prediction.
The relationship between $\rvz_{\phi_t}, \rvv_{t}, \alpha_{t}, \sigma_{t}, \rvx, \rvepsilon$ is visualized in~\cref{fig:phi}.

\section{More Empirical Details}

\subsection{Detailed Algorithm}
\label{sec:detailed_alg}
Due to space limitations, we omitted some implementation details in the main body, but we provided a detailed version of the OMS based on DDIM sampling in Alg.~\ref{alg1}. This example implementation utilizes $\rvv$-prediction for the OMS and $\eps$-prediction for the pre-trained model.

\begin{algorithm}[htb!]
\SetAlgoLined
\KwRequire{Pre-trained Diffusion Pipeline with a model $\theta$ to perform $\epsilon$-prediction.}
\KwRequire{One More Step module $\psi(\cdot)$}
\KwInput{OMS Text Prompt $\mathcal{C}_\psi$, OMS CFG weight $\omega_\psi$}
\KwInput{Text Prompt $\mathcal{C}_\theta$, Guidance weight $\omega_\theta$, Eta $\sigma$}
{\color{RoyalBlue}{\# Introduce One More Step}} \\
$\rvz \sim \mathcal{N}(0, \mathbf{I})$ \;
$\rvx_T^{\mathcal{S}} \sim \mathcal{N}(0, \mathbf{I}$)\;
{\color{RoyalBlue}{\# Classifier Free Guidance at One More Step Phase}} \\
\uIf{$\omega_\psi>1$}{
$\tilde{\rvx}_0^{\mathcal{S}} = -\psi_\text{cfg}(\rvx_T^{\mathcal{S}}, \mathcal{C}_\psi, \emptyset, \omega_\psi)$ \;
}
\Else{
$\tilde{\rvx}_0^{\mathcal{S}} = -\psi(\rvx_T^{\mathcal{S}}, \mathcal{C}_\psi)$ \;
}
$\tilde{\rvx}_{T}^{\mathcal{T}} =  \sqrt{\bar{\alpha}_T^{\mathcal{T}}} \tilde{\rvx}_0^{\mathcal{S}} + \sqrt{1-\bar{\alpha}_T^{\mathcal{T}} - \sigma^2} \rvx_T^{\mathcal{S}} + \sigma \rvz$ \;
{\color{RoyalBlue}{\# Sampling from Pre-trained Diffusion Model}} \\
\For{$t = T, \dots, 1$}{
$\rvz \sim \mathcal{N}(0, \mathbf{I})$ if $t>1$, else $\rvz=0$\;
\uIf{t=T}{
    \uIf{$\omega_\theta>1$}{
    $\tilde\eps_T = \theta_\text{cfg}(\tilde\rvx_T^{\mathcal{T}}, \mathcal{C}_\theta, \emptyset, \omega_\theta)$ \;
    }
    \Else{
    $\tilde\eps_T = \theta (\tilde\rvx_t^{\mathcal{T}}, \mathcal{C}_\theta)$ \;
    }
$\tilde{\rvx}_{T-1} = \sqrt{\bar{\alpha}_{T-1}}\left(\frac{\tilde{\rvx}_{T}^{\mathcal{T}}-\sqrt{1-\bar{\alpha}_T^{\mathcal{T}}}\tilde\eps_T}{\sqrt{\bar\alpha_T^{\mathcal{T}}}}\right) +  \sqrt{1-\bar{\alpha}_{T-1} - \sigma^2}\tilde\eps_T +\sigma\rvz $ \;
} 
\Else{
    \uIf{$\omega_\theta>1$}{
    $\tilde\eps_t = \theta_\text{cfg}(\tilde\rvx_t, \mathcal{C}_\theta, \emptyset, \omega_\theta)$ \;
    }
    \Else{
    $\tilde\eps_t = \theta (\rvx_t^{\mathcal{T}}, \mathcal{C}_\theta)$ \;
    }
   $\tilde{\rvx}_{t-1} = \sqrt{\bar{\alpha}_{t-1}}\left(\frac{\tilde{\rvx}_{t}-\sqrt{1-\bar{\alpha}_t} \tilde\eps_t}{\sqrt{\bar\alpha_t}}\right) +  \sqrt{1-\bar{\alpha}_{t-1} - \sigma^2}\tilde\eps_t +\sigma\rvz $ 
}  
}
\Return{$\tilde{\rvx}_0$}
\caption{DDIM Sampling with OMS}
\label{alg1}
\end{algorithm}

The derivation related to prediction of $\tilde{\rvx}_{T}^{\mathcal{T}}$ in Eq.~\ref{omsddim} can be obtained from Eq.12 in~\cite{song2020denoising}.
Given $\rvx_t$, one can generate $\rvx_0$:

\begin{equation}
    \tilde{\rvx}_{t-1} = \sqrt{\bar{\alpha}_{t-1}}\left(\frac{\tilde{\rvx}_{t}-\sqrt{\bar{\alpha}_t} \tilde\eps_t}{\sqrt{\bar\alpha_t}}\right) +  \sqrt{1-\bar{\alpha}_{t-1} - \sigma_t^2}\tilde\eps_t +\sigma_t\rvz, 
    \label{ddim_org}
\end{equation}
where $\tilde\rvx_0^t$ is parameterised by $\frac{\tilde{\rvx}_{t}-\sqrt{\bar{\alpha}_t} \tilde\eps_t}{\sqrt{\bar\alpha_t}}$. In OMS phase, $\bar{\alpha}_T^{\mathcal{S}}=0$ and $\bar{\alpha}_{T-1}^{\mathcal{S}} = \bar{\alpha}_{T}^{\mathcal{T}}$. According to Eq.~\ref{v_mu}, the OMS module $\psi(\cdot)$ directly predict the direction $\rvv$ of the data, which is equal to $-\tilde{\rvx}_0^{\mathcal{S}}$:

\begin{equation}
    \tilde{\rvx}_0^{\mathcal{S}} :=  - \rvv_{\psi}(\rvx_T^{\mathcal{S}}, \mathcal{C}).
\end{equation}
Applying these conditions to Eq.~\ref{ddim_org} yields the following:

\begin{equation}
    \tilde{\rvx}_{T}^{\mathcal{T}} =  \sqrt{\bar{\alpha}_T^{\mathcal{T}}} \tilde{\rvx}_0^{\mathcal{S}} + \sqrt{1-\bar{\alpha}_T^{\mathcal{T}} - \sigma^2} \rvx_T^{\mathcal{S}} + \sigma \rvz
\end{equation}

\subsection{Additional Comments}
\label{adco}

\paragraph{Alternative training targets for OMS} As we discussed in~\ref{ptt}, the objective of $\rvv$-prediction at SNR=0 scenario is exactly the same as negative $\rvx_0$-prediction. Thus we can also train the OMS module under the L2 loss between $\Vert \rvx_0 - \tilde \rvx_0 \Vert_2^2$, where the OMS module directly predict $\tilde \rvx_0 = \psi (\rvx_T^{\mathcal{S}}, \mathcal{C})$.

\paragraph{Reasons behind versatility} The key point is revealed in Eq.~\ref{omsddim}. The target prediction of OMS module is only focused on the conditional mean value $\tilde \rvx_0$, which is only related to the training data. $\rvx_T^{\mathcal{S}}$ is directly sampled from normal distribution, which is independent. Only $\bar\alpha_T$ is unique to other pre-defined diffusion pipelines, but it is non-parametric. Therefore, given an $\rvx_T^{\mathcal{S}}$ and an OMS module $\psi$, we can calculate any $\rvx_T^{\mathcal{T}}$ that aligns with the pre-trained model schedule according to Eq.~\ref{omsddim}.

\paragraph{Consistent generation} Additionally, our study demonstrates that the OMS can significantly enhance the coherence and continuity between the generated images, which aligns with the discoveries presented in recent research~\cite{girdhar2023emu} to improve the coherence between frames in the video generation process.

\subsection{Implementation Details}
\label{appe_imple}

\paragraph{Dataset}
The proposed OMS module and its variants were trained on the LAION 2B dataset~\cite{schuhmann2022laion} without employing any specific filtering operation. All the training images are first resized to 512 pixels by the shorter side and then randomly cropped to dimensions of 512 × 512, along with a random flip. Notably, for the model trained on the pretrained SDXL, we utilize a resolution of 1024.
Additionally, we conducted experiments on LAION-HR images with an aesthetic score greater than 5.8. However, we observed that the high-quality dataset did not yield any improvement. This suggests that the effectiveness of our model is independent of data quality, as OMS predicts the mean of training data conditioned on the prompt.

\paragraph{OMS scale variants}
We experiment with OMS modules at three different scales, and the detailed settings for each variants are shown in Table~\ref{tab:variants}. 
Combining these with three different text encoders results in a total of nine OMS modules with different parameters.
As demonstrated in Table~\ref{tab:size_text}, we found that OMS is not sensitive to the number of parameters and the choice of text encoder used to extract text embeddings for the OMS network.
\begin{table}[]
\centering
\resizebox{\columnwidth}{!}{
\begin{tabular}{llll}
\toprule
Model  &OMS-S  &OMS-B  &OMS-L  \\ \midrule
Layer num. & 2 & 2 & 2 \\
Transformer blocks & 1 & 1 & 1 \\
Channels & [32, 64, 64] & [160, 320, 640] & [320, 640, 1280, 1280] \\
Attention heads & [2, 4, 4] & 8 & [5, 10, 20, 20] \\
Cross Attn dim. & 768/1024/4096 & 768/1024/4096 & 768/1024/4096  \\
\# of OMS params & 3.3M/3.7M/8.1M & 151M/154M/187M & 831M/838M/915M  \\ \bottomrule
\end{tabular}
}
\caption{Model scaling variants of OMS.}
\label{tab:variants}
\end{table}

\begin{table}
\centering
\begin{subtable}[t]{\linewidth}
\begin{tabular}{llll}
\toprule
OMS Scale &CLIP ViT-L  &OpenCLIP ViT-H  &T5-XXL  \\ \midrule
OMS-S & 45.87 & 45.30 & 45.35 \\
OMS-B & 46.85 & 45.74 & 45.77 \\
OMS-L & 46.68 & 45.65 & 45.19  \\ \bottomrule
\end{tabular}
\caption{ImageReward results among different OMS scales and text encoders}
\end{subtable}
\hfill
\begin{subtable}[t]{\linewidth}
\begin{tabular}{llll}
\toprule
OMS Scale &CLIP ViT-L  &OpenCLIP ViT-H  &T5-XXL  \\ \midrule
OMS-S & 21.82 & 21.82 & 21.80 \\
OMS-B & 21.83 & 21.82 & 21.81 \\
OMS-L & 21.82 & 21.82 & 21.80  \\ \bottomrule
\end{tabular}
\caption{PickScore results among different OMS scales and text encoders}
\end{subtable}

\caption{Experiment results among different OMS scales and text encoders on pre-trained SD2.1.}
\label{tab:size_text}
\end{table}

\paragraph{Hyper-parameters}
In our experiments, we employed the AdamW optimizer with $\beta_1 = 0.9$, $\beta_2 = 0.999$, and a weight decay of 0.01. The batch size and learning rate are adjusted based on the model scale, text encoder, and pre-trained model, as detailed in Tab.~\ref{tab:hyper}. Notably, our observations indicate that our model consistently converges within a relatively low number of iterations, typically around 2,000 iterations being sufficient.

\paragraph{Hardware and speed}
All our models were trained using eight 80G A800 units, and the training speeds are provided in Tab.~\ref{tab:hyper}. It is evident that our model was trained with high efficiency, with OMS-S using CLIP ViT-L requiring only about an hour for training.

\begin{table}
\centering
\resizebox{\columnwidth}{!}{
\begin{tabular}{llll}
\toprule
Model  &Batch size  &Learning rate  &Training time  \\ \midrule
OMS-S/CLIP (SD2.1) & 512 & 5.0e-5 & 1.21h \\
OMS-B/CLIP (SD2.1) & 512 & 5.0e-5 & 1.37h  \\
OMS-L/CLIP (SD2.1) & 512 & 5.0e-5 & 1.98h  \\
OMS-S/OpenCLIP (SD2.1) & 512 & 5.0e-5 & 1.21h \\
OMS-B/OpenCLIP (SD2.1) & 512 & 5.0e-5 & 1.37h  \\
OMS-L/OpenCLIP (SD2.1) & 512 & 5.0e-5 & 2.00h  \\
OMS-S/T5 (SD2.1) & 256 & 3.5e-5 & 1.49h \\
OMS-B/T5 (SD2.1) & 256 & 3.5e-5 & 1.56h  \\
OMS-L/T5 (SD2.1) & 256 & 3.5e-5 & 2.07h  \\
OMS-S/OpenCLIP (SDXL) & 128 & 2.5e-5 & 1.46h  \\
OMS-B/OpenCLIP (SDXL) & 128 & 2.5e-5 & 1.65h  \\
OMS-L/OpenCLIP (SDXL) & 128 & 2.5e-5 & 2.68h  \\
\bottomrule
\end{tabular}
}
\caption{Distinct hyper-parameters and training speed on different model. All models are trained for 2k iterations using 8 80G A800.}
\label{tab:hyper}
\end{table}

\subsection{OMS Versatility and VAE Latents Domain}
\label{vae_con}

The output of the OMS model is related to the training data of the diffusion phase. If the diffusion model is trained in the image domain, then our image domain-based OMS can be widely applied to these pre-trained models. However, the more popular LDM model has a VAE as the first stage that compresses the pixel domain into a latent space. For different LDM models, their latent spaces are not identical. In such cases, the training data for OMS is actually the latent compressed by the VAE Encoder. Therefore, our OMS model is versatile for pre-trained LDM models within the same VAE latent domain, \eg, SD1.5, SD2.1 and LCM. 

\begin{figure}[ht]
    \centering
    \begin{subfigure}[b]{\linewidth}

        \includegraphics[width=\linewidth]{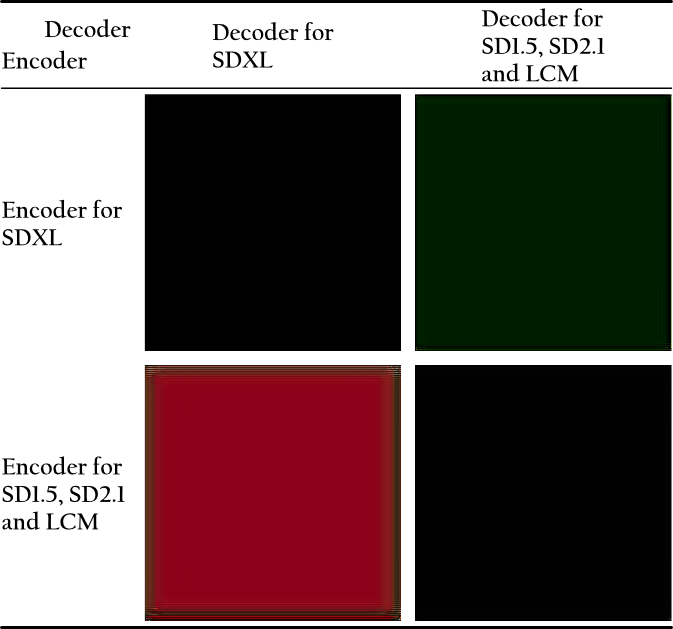}
            \caption{Encode and Decode Black Image with Different VAEs}
    \end{subfigure}
    \begin{subfigure}[b]{\linewidth}

        \includegraphics[width=\linewidth]{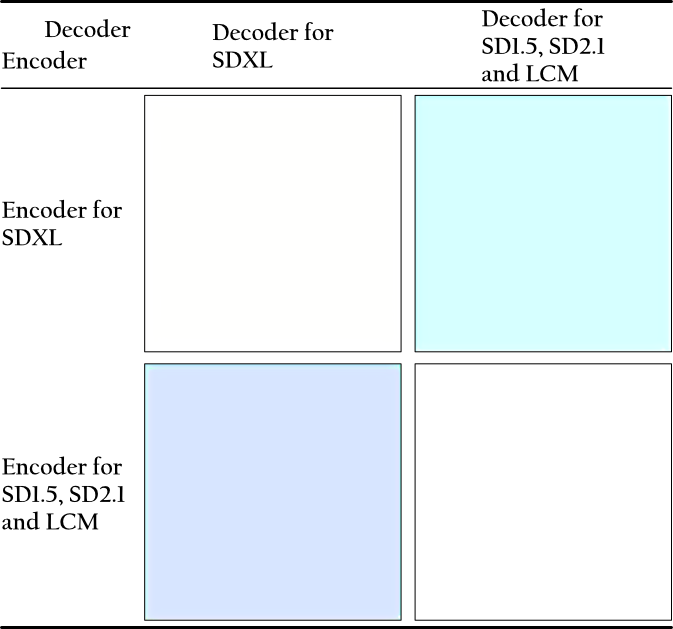}
            \caption{Encode and Decode White Image with Different VAEs}
    \end{subfigure}
    \caption{The offset in compression and reconstruction of different series of VAEs.}
    \label{fig:vae}
\end{figure}

Our analysis reveals that the VAEs in SD1.5, SD2.1, and LCM exhibit a parameter discrepancy of less than 1e-4 and are capable of accurately restoring images. Therefore, we consider that these three are trained diffusion models in the same latent domain and can share the same OMS module. However, for SDXL, our experiments found significant deviations in the reconstruction process, especially in more extreme cases as shown in~\cref{fig:vae}. Therefore, the OMS module for SDXL needs to be trained separately. But it can still be compatible with other models in the community based on SDXL.

\begin{figure}[ht]
    \centering
    \begin{subfigure}[b]{\linewidth}
        \includegraphics[width=\linewidth]{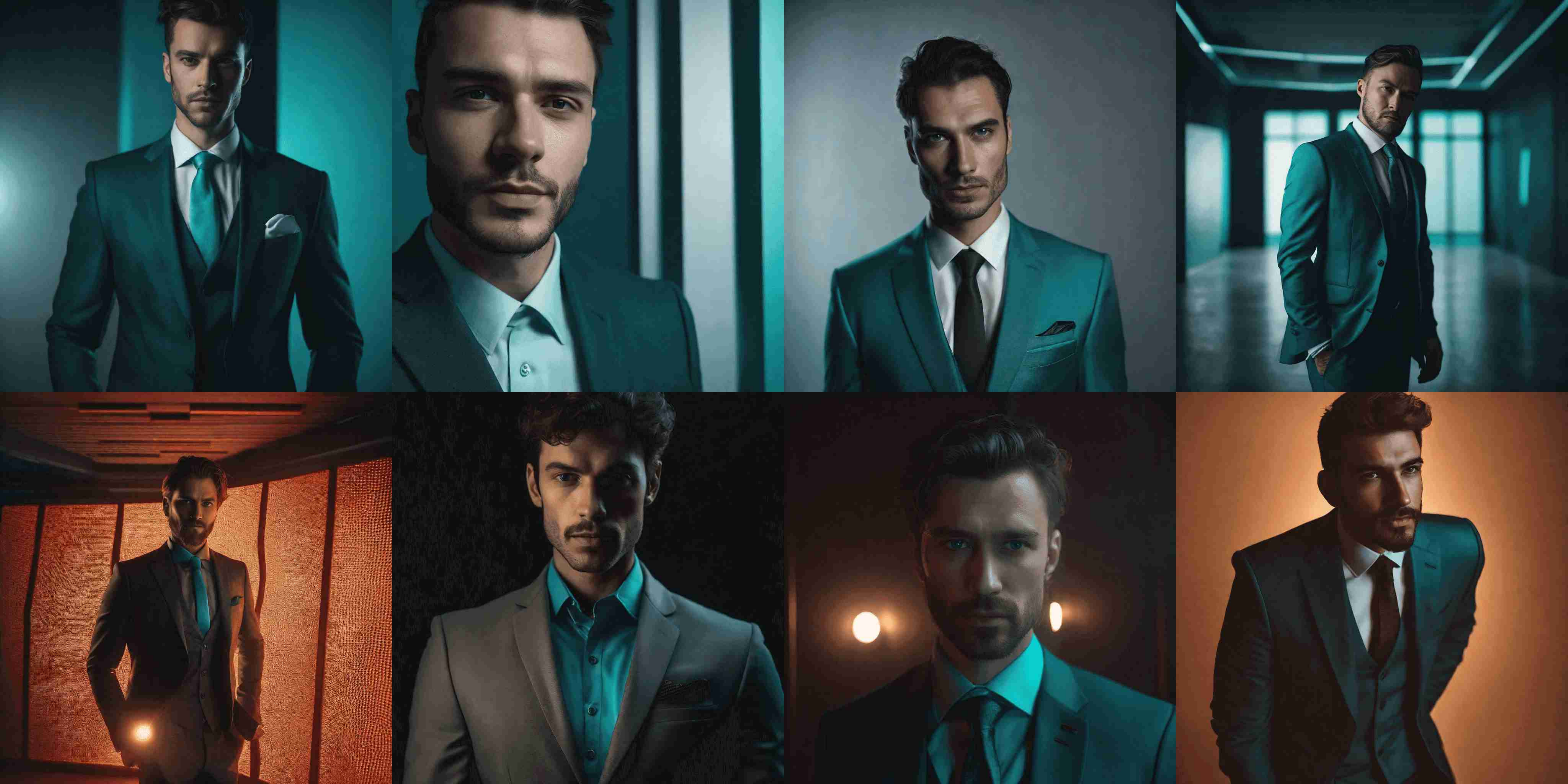}
        \caption{\textbf{Close-up portrait of a man wearing suit posing in a dark studio, rim lighting, teal hue, octane, unreal}}
    \end{subfigure}
    \begin{subfigure}[b]{\linewidth}
        \includegraphics[width=\linewidth]{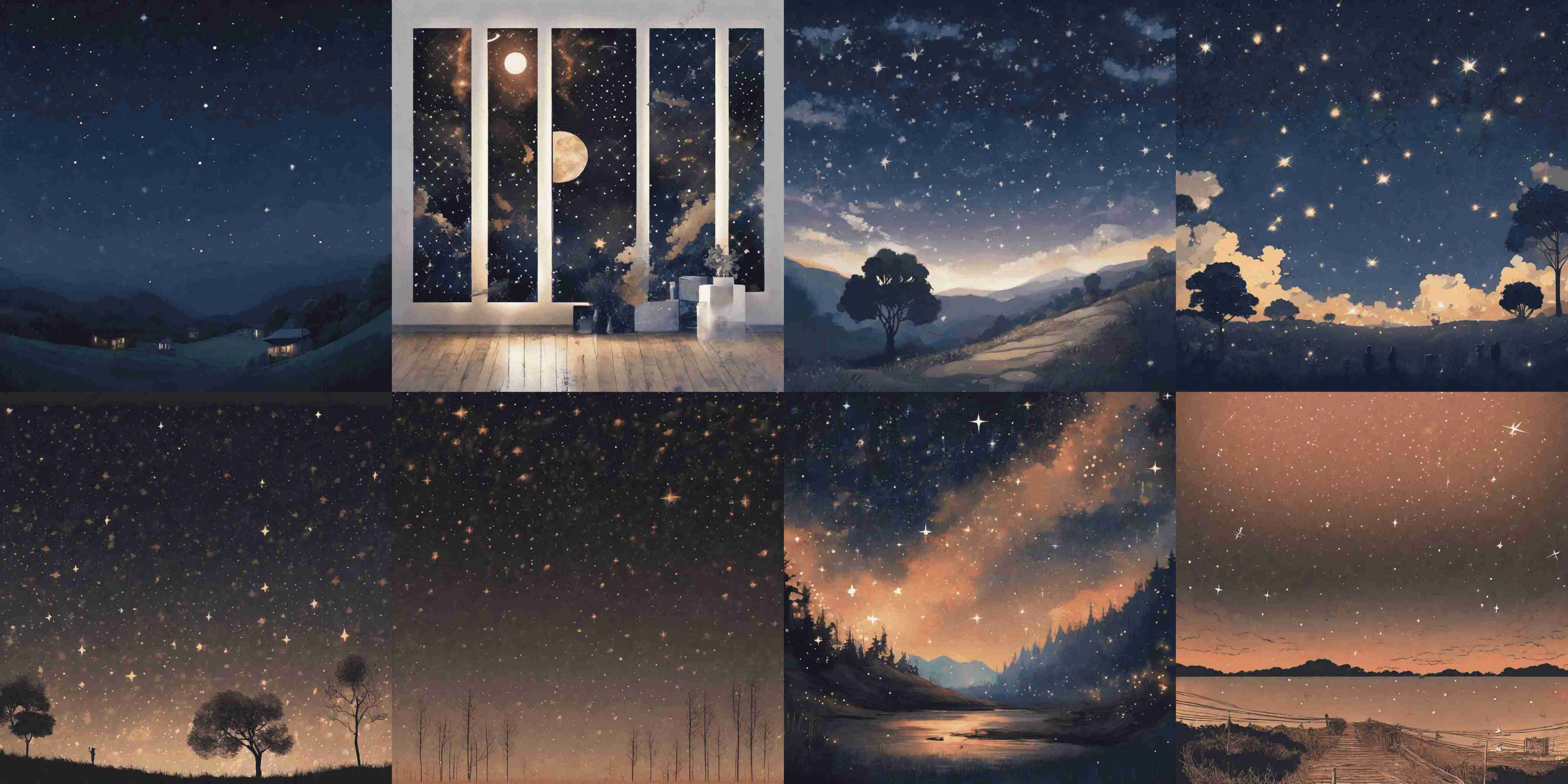}
        \caption{\textbf{A starry sky}}
    \end{subfigure}
    \caption{Examples of distortion due to incompatible VAEs. Use the OMS model trained on SD1.5 VAE to forcibly conduct inference on SDXL base model. The upper layer of each subfigure shows the results sampled using the original model, while the lower layer shows the results of inference using the biased OMS model.}
    \label{fig:failure}
\end{figure}

If we forcibly use the OMS trained with the VAE of the SD1.5 series on the base model of SDXL, severe color distortion will occur whether we employ latents with unit variance. We demonstrate some practical distortion case with the rescaled unit variance space in~\cref{fig:failure}. The observed color shift aligns with the effect shown in~\cref{fig:vae}, \eg, Black $\rightarrow$ Red.

\begin{figure*}[ht!]
    \centering
    \begin{subfigure}[b]{0.45\linewidth}
        \includegraphics[width=\linewidth]{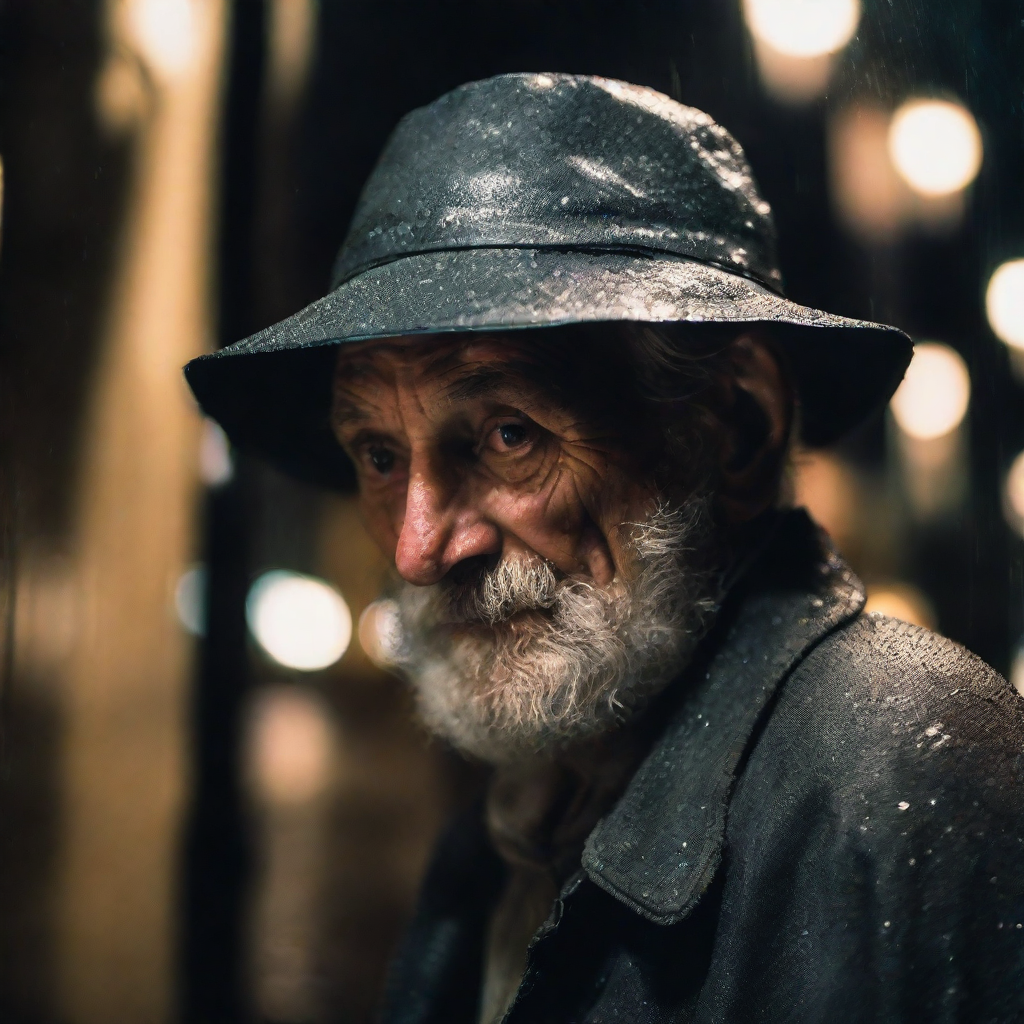}
        \caption{\textbf{close-up photography of old man standing in the rain at night, in a street lit by lamps, leica 35mm summilux}, SDXL with LCM-LoRA, LCM Scheduler with 4 Steps. CFG weight is 1 (no CFG). Mean value is 0.24.}
    \end{subfigure}
        \begin{subfigure}[b]{0.45\linewidth}
        \includegraphics[width=\linewidth]{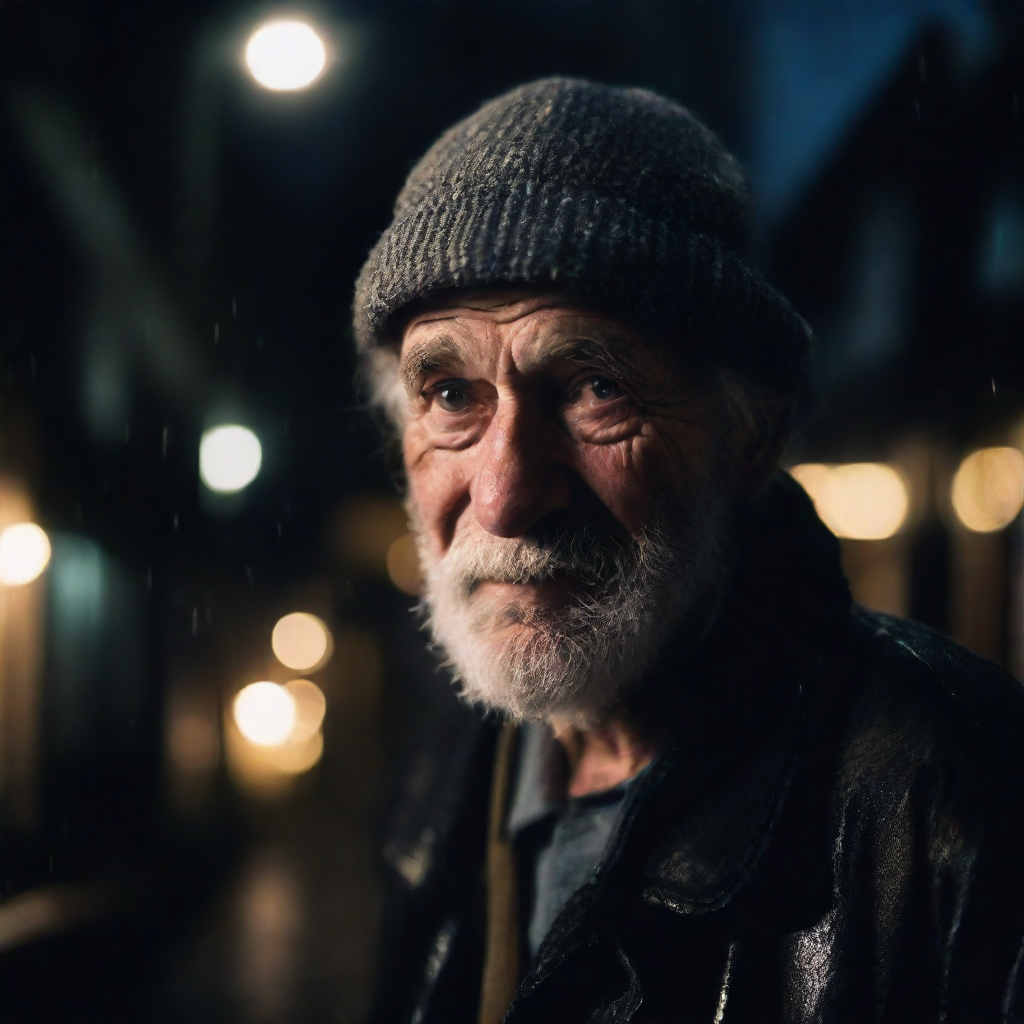}
        \caption{\textbf{close-up photography of old man standing in the rain at night, in a street lit by lamps, leica 35mm summilux}, SDXL with LCM-LoRA, LCM Scheduler with 4 + 1 (OMS) Steps. Base model CFG is 1 and OMS CFG is 2. Mean value is \textbf{0.14}.}
    \end{subfigure}
    \caption{LCM-LoRA on SDXL for the reproduced result.}
    \label{fig:lcm_lora_rep}
\end{figure*}

\section{More Experimental Results}
\label{sec:extre}

\subsection{LoRA and Community Models}

\begin{figure*}[ht]
    \centering
    \begin{subfigure}[b]{0.75\linewidth}
        \includegraphics[width=\linewidth]{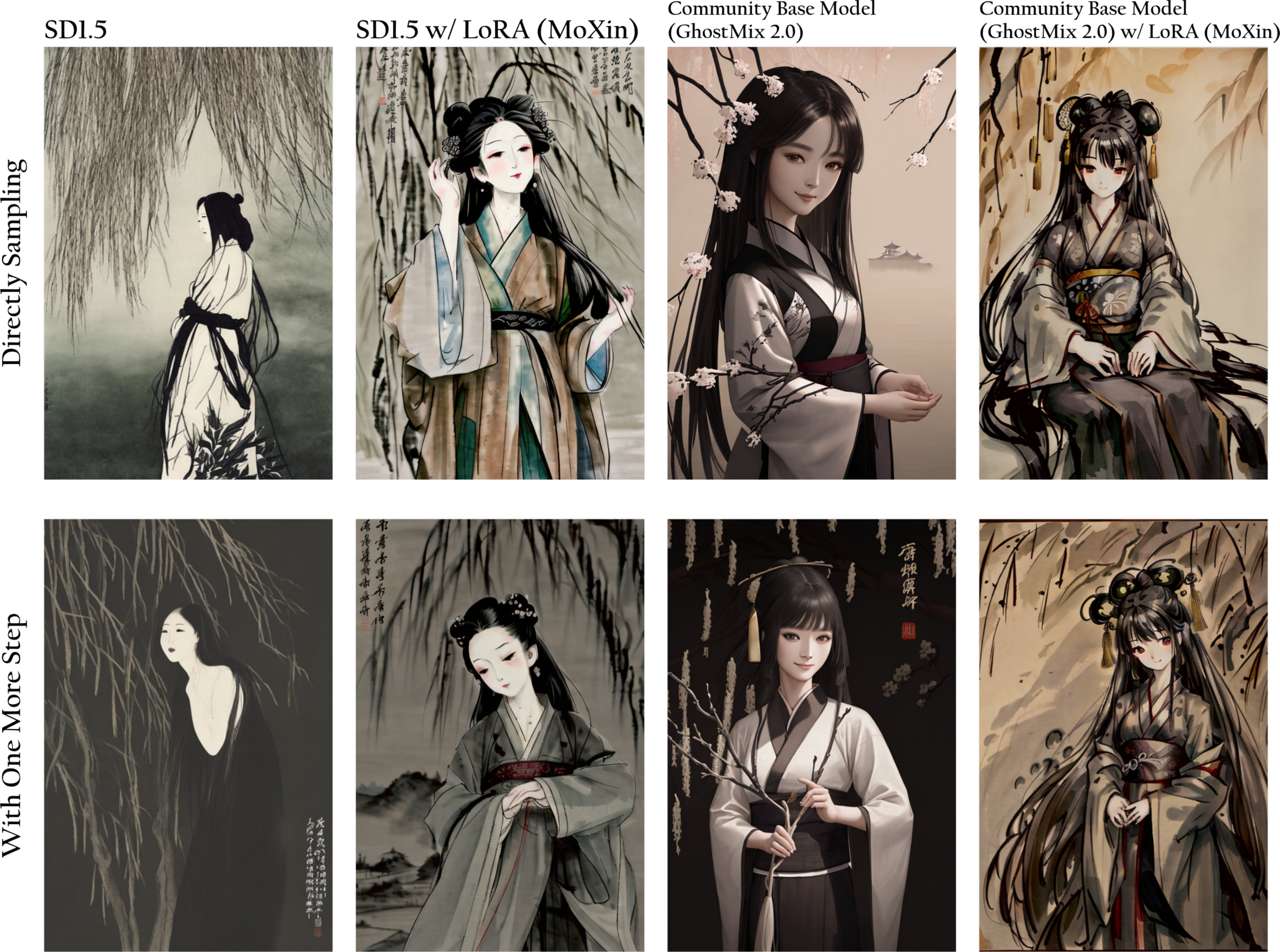}
        \caption{\textbf{portrait of a woman standing , willow branches, masterpiece, best quality, traditional chinese ink painting, modelshoot style, peaceful, smile, looking at viewer, wearing long hanfu, song, willow tree in background, wuchangshuo, high contrast, in dark, black}}
    \end{subfigure}
    \begin{subfigure}[b]{0.75\linewidth}
        \includegraphics[width=\linewidth]{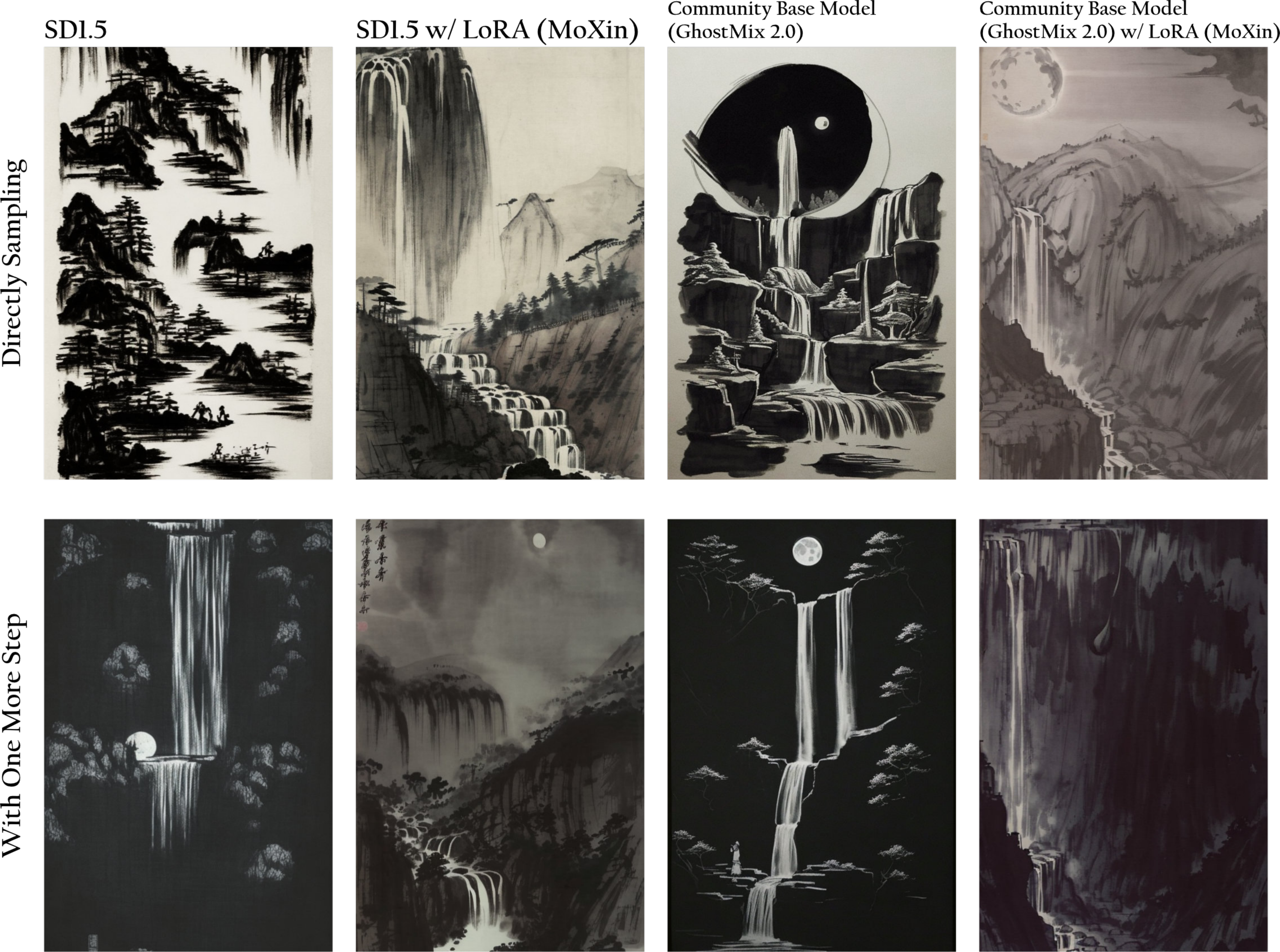}
        \caption{\textbf{The moon and the waterfalls, night, traditional chinese ink painting, modelshoot style, masterpiece, high contrast, in dark, black}}
    \end{subfigure}
    \caption{Examples of SD1.5, Community Base Model \textit{GhostMix} and LoRA \textit{MoXin} with OMS leading to darker images.}
    \label{fig:loras_b}
\end{figure*}

\begin{figure*}[ht]
    \centering
    \begin{subfigure}[b]{0.75\linewidth}
        \includegraphics[width=\linewidth]{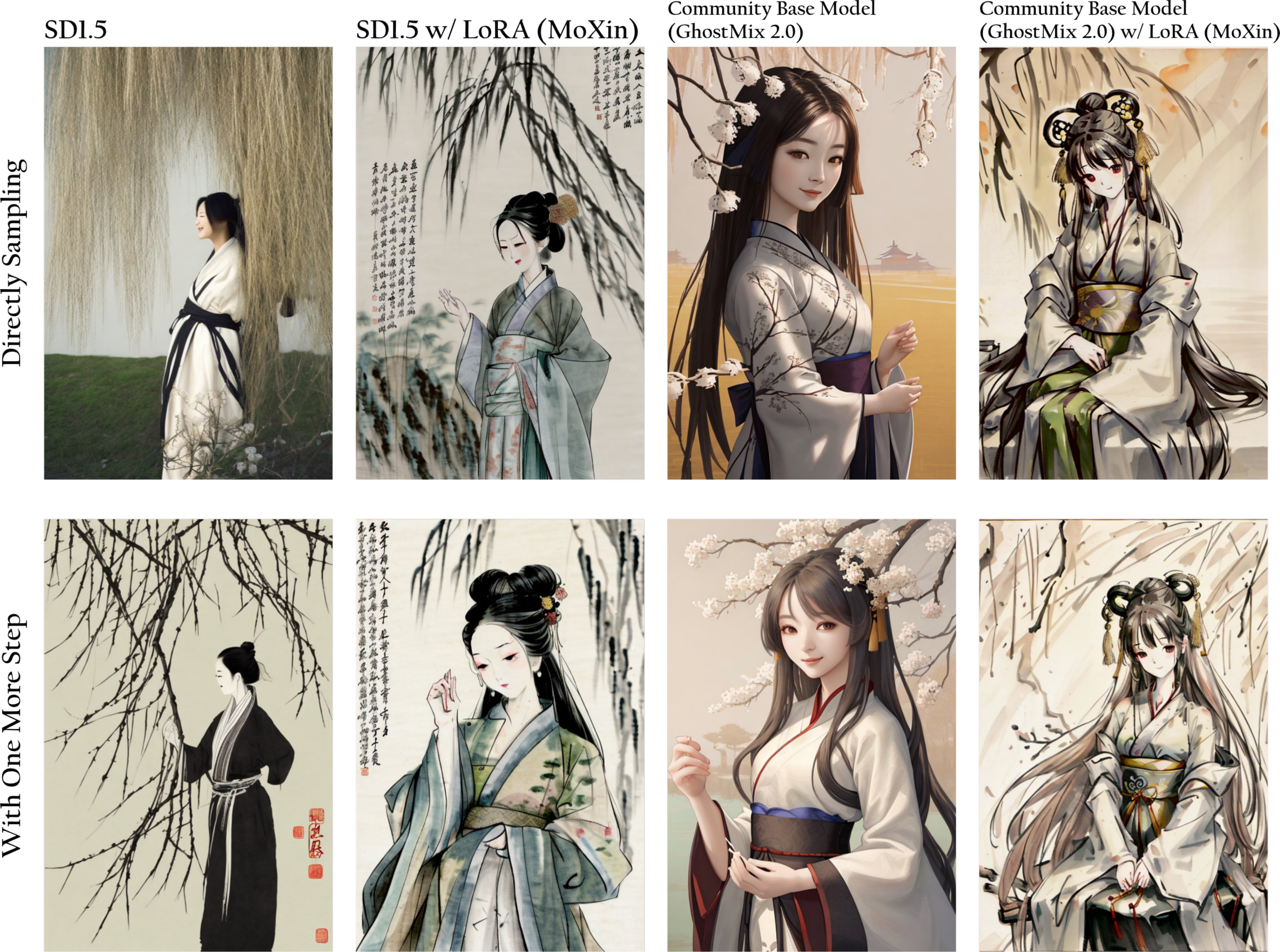}
        \caption{\textbf{portrait of a woman standing , willow branches, masterpiece, best quality, traditional chinese ink painting, modelshoot style, peaceful, smile, looking at viewer, wearing long hanfu, song, willow tree in background, wuchangshuo, high contrast, in sunshine, white}}
    \end{subfigure}
    \begin{subfigure}[b]{0.75\linewidth}
        \includegraphics[width=\linewidth]{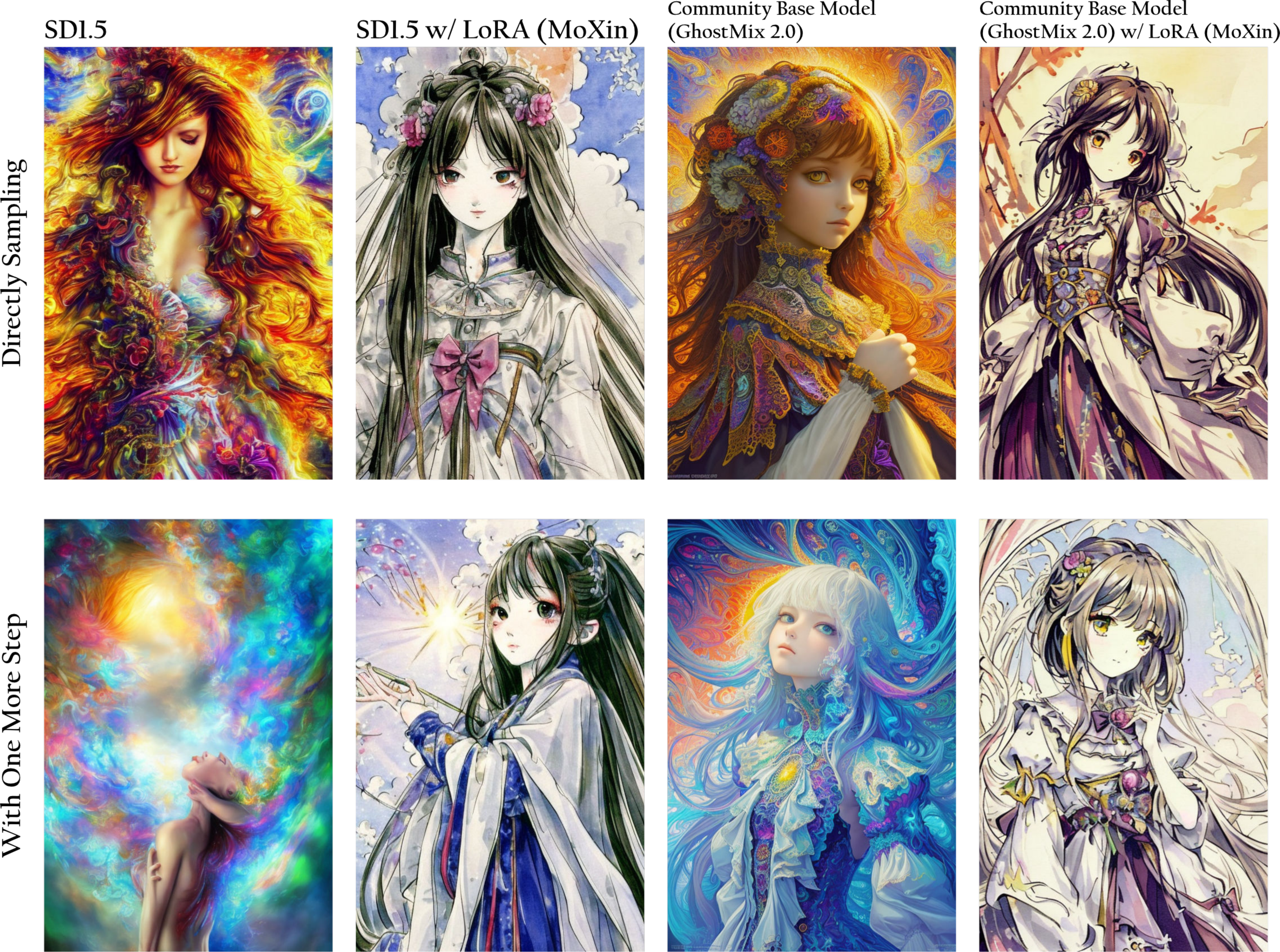}
        \caption{\textbf{(masterpiece, top quality, best quality, official art, beautiful and aesthetic:1.2), (1girl), extreme detailed,(fractal art:1.3),colorful,highest detailed, high contrast, in sunshine, white}}
    \end{subfigure}
    \caption{Examples of SD1.5, Community Base Model \textit{GhostMix} and LoRA \textit{MoXin} with OMS leading to brighter images.}
    \label{fig:loras_w}
\end{figure*}

In this experiment, we selected a popular community model \textit{GhostMix 2.0 BakedVAE}~\footnote{\textit{GhostMix} can be found at \texttt{https://civitai.com/models/36520}} and a LoRA \textit{MoXin 1.0}~\footnote{\textit{MoXin} can be found at \texttt{https://civitai.com/models/12597}}. In~\cref{fig:loras_b}~\&~\cref{fig:loras_w}, we see that the OMS module can be applied to many scenarios with obvious effects. LoRA scale is set as 0.75 in the experiments. We encourage readers to adopt our method in a variety of well-established open-source models to enhance the light and shadow effects in generated images.

We also do some experiment on LCM-LoRA~\cite{luo2023lcm} with SDXL for fast inference. The OMS module is the same as we used for SDXL.

\subsection{Additional Results}

\begin{figure*}[ht]
    \centering
    \begin{subfigure}[b]{0.7\linewidth}
        \includegraphics[width=\linewidth]{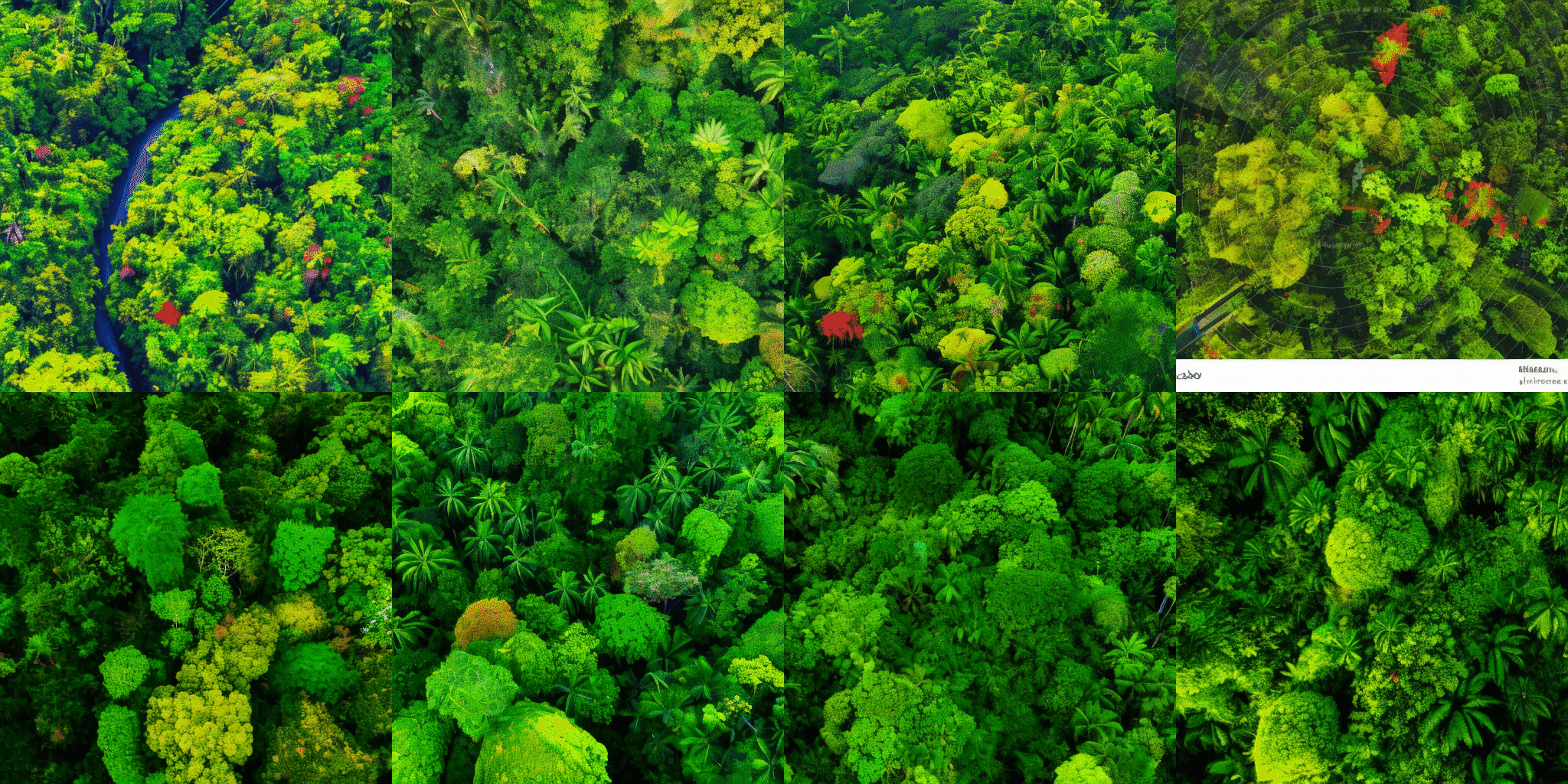}
        \caption{Aerial view of a vibrant tropical rainforest, filled with lively green vegetation and colorful flowers, sunlight piercing through the canopy, high contrast, vivid colors}
    \end{subfigure}
    
    \begin{subfigure}[b]{0.7\linewidth}
        \includegraphics[width=\linewidth]{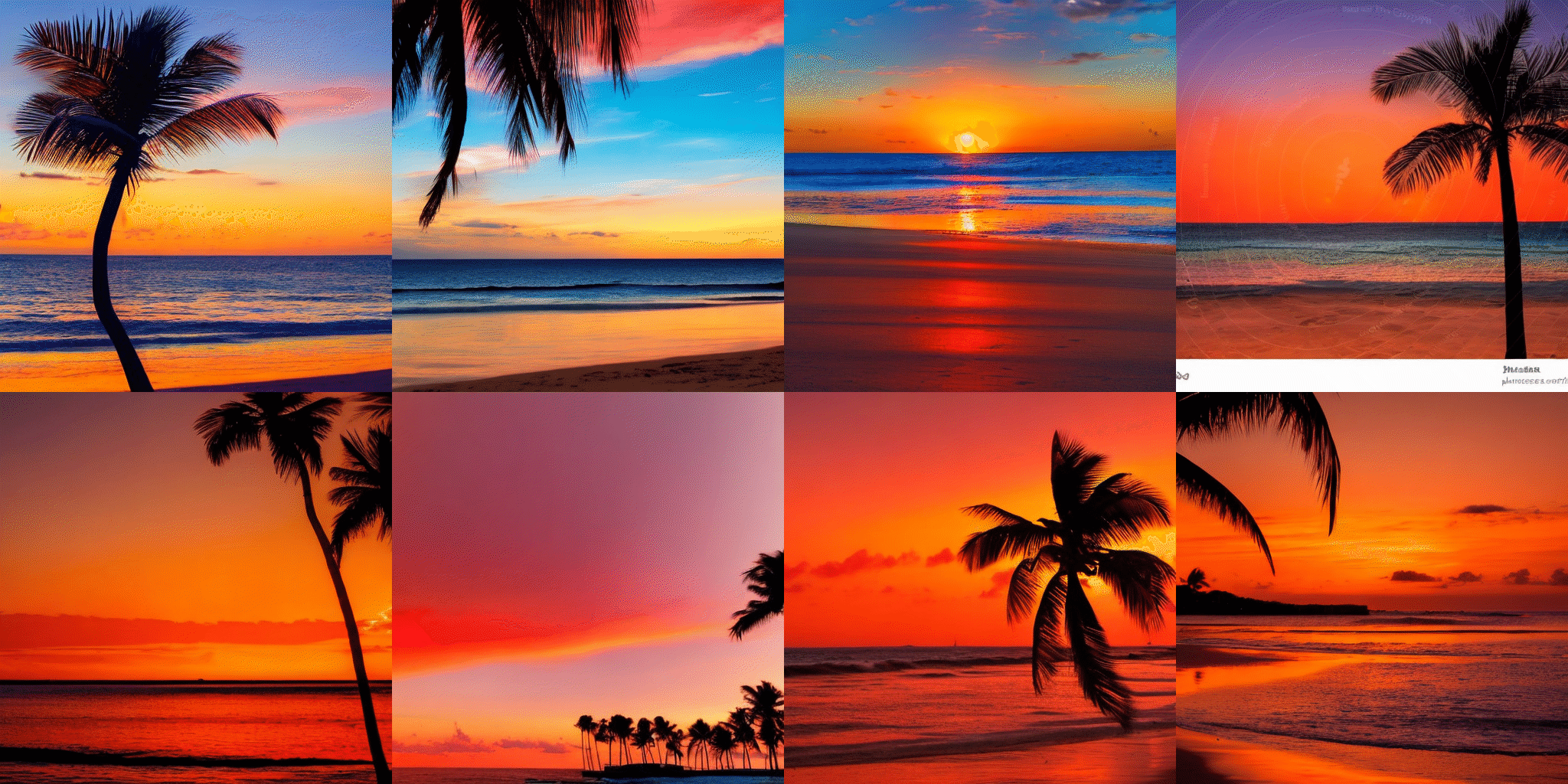}
        \caption{Tropical beach at sunset, the sky in splendid shades of orange and red, the sea reflecting the sun's afterglow, clear silhouettes of palm trees on the beach, high contrast, vivid colors}
    \end{subfigure}
    
    \begin{subfigure}[b]{0.7\linewidth}
        \includegraphics[width=\linewidth]{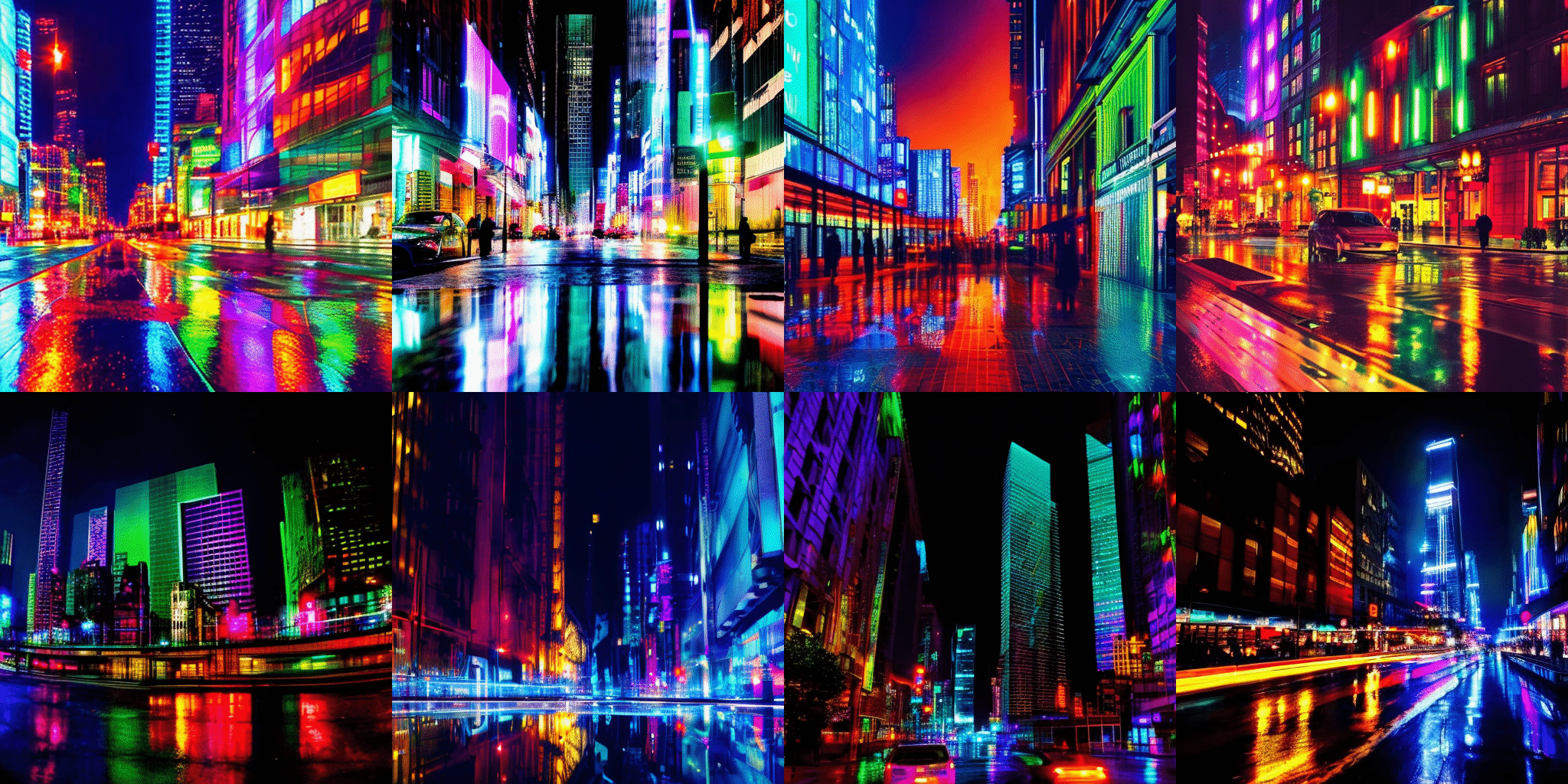}
        \caption{A cityscape at night with neon lights reflecting off wet streets, towering skyscrapers illuminated in a kaleidoscope of colors, high contrast between the bright lights and dark shadows}
    \end{subfigure}

    \caption{Additional Samples from SD1.5, top row from original model and bottom row with OMS.}
    \label{fig:sd15}
\end{figure*}

\begin{figure*}[ht]
    \centering
    \begin{subfigure}[b]{0.7\linewidth}
        \includegraphics[width=\linewidth]{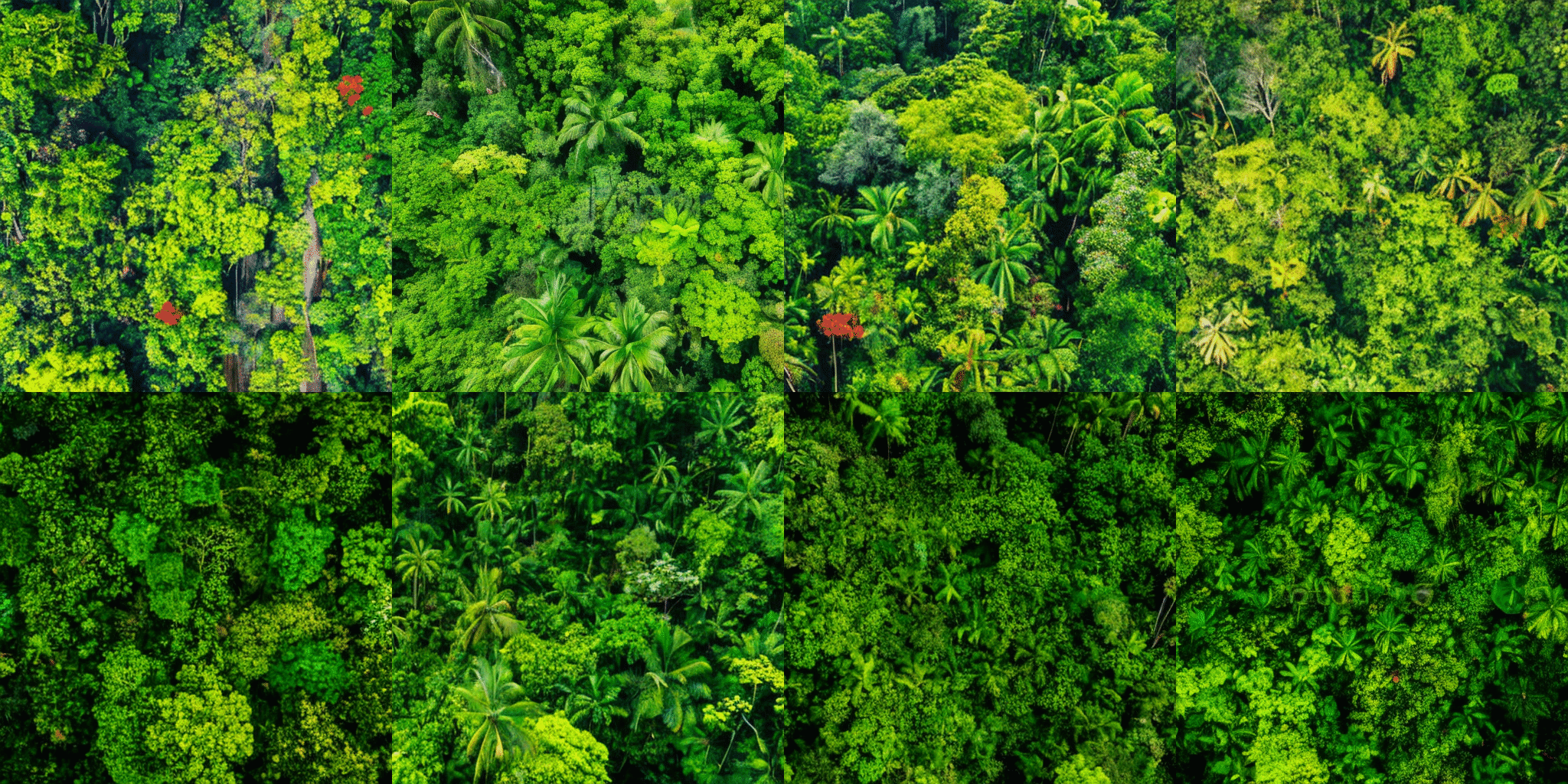}
        \caption{Aerial view of a vibrant tropical rainforest, filled with lively green vegetation and colorful flowers, sunlight piercing through the canopy, high contrast, vivid colors}
    \end{subfigure}
    
    \begin{subfigure}[b]{0.7\linewidth}
        \includegraphics[width=\linewidth]{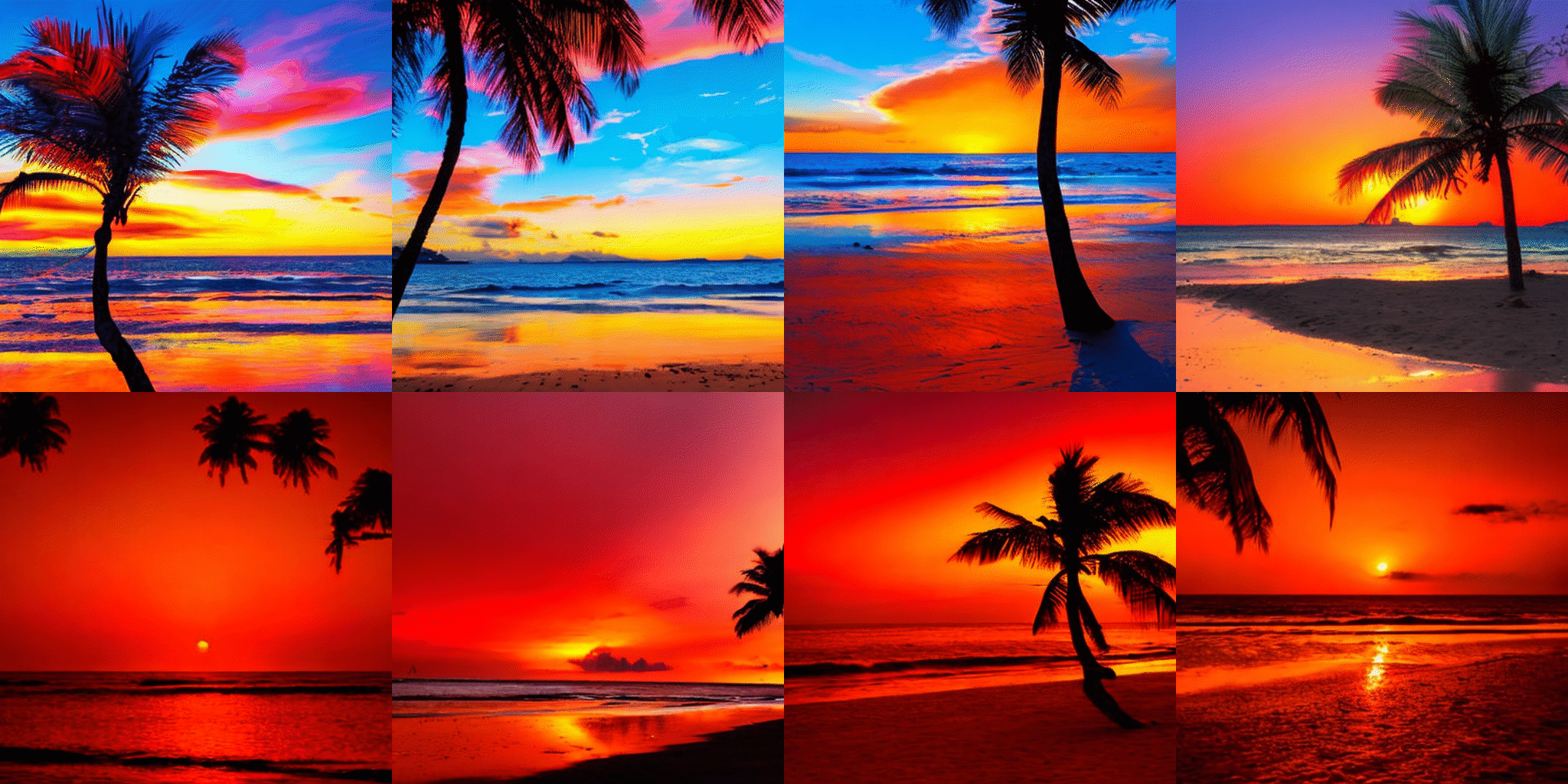}
        \caption{Tropical beach at sunset, the sky in splendid shades of orange and red, the sea reflecting the sun's afterglow, clear silhouettes of palm trees on the beach, high contrast, vivid colors}
    \end{subfigure}
    
    \begin{subfigure}[b]{0.7\linewidth}
        \includegraphics[width=\linewidth]{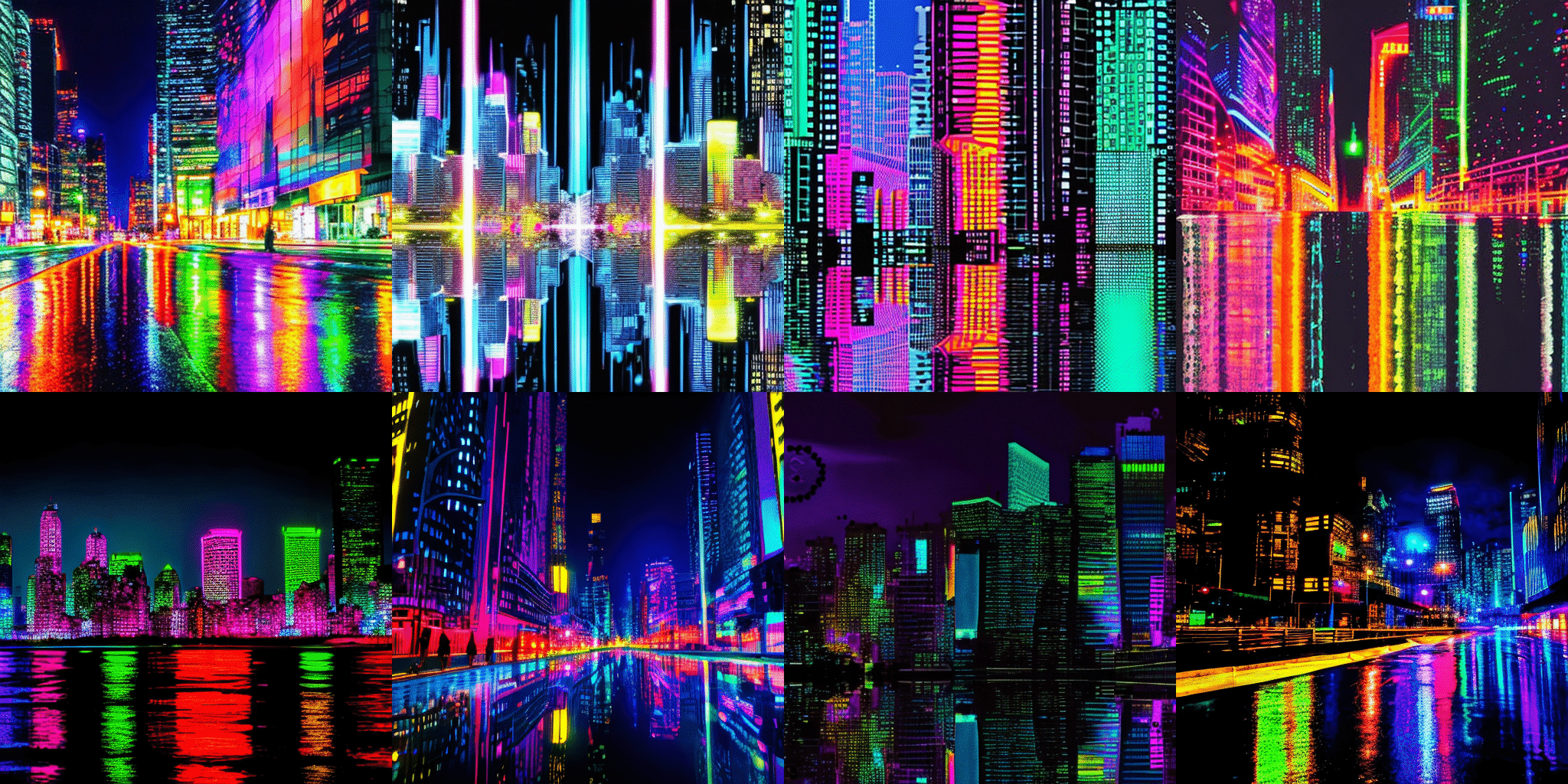}
        \caption{A cityscape at night with neon lights reflecting off wet streets, towering skyscrapers illuminated in a kaleidoscope of colors, high contrast between the bright lights and dark shadows}
    \end{subfigure}

    \caption{Additional Samples from SD2.1, top row from original model and bottom row with OMS.}
    \label{fig:sd21}
\end{figure*}

\begin{figure*}[ht]
    \centering
    \begin{subfigure}[b]{0.7\linewidth}
        \includegraphics[width=\linewidth]{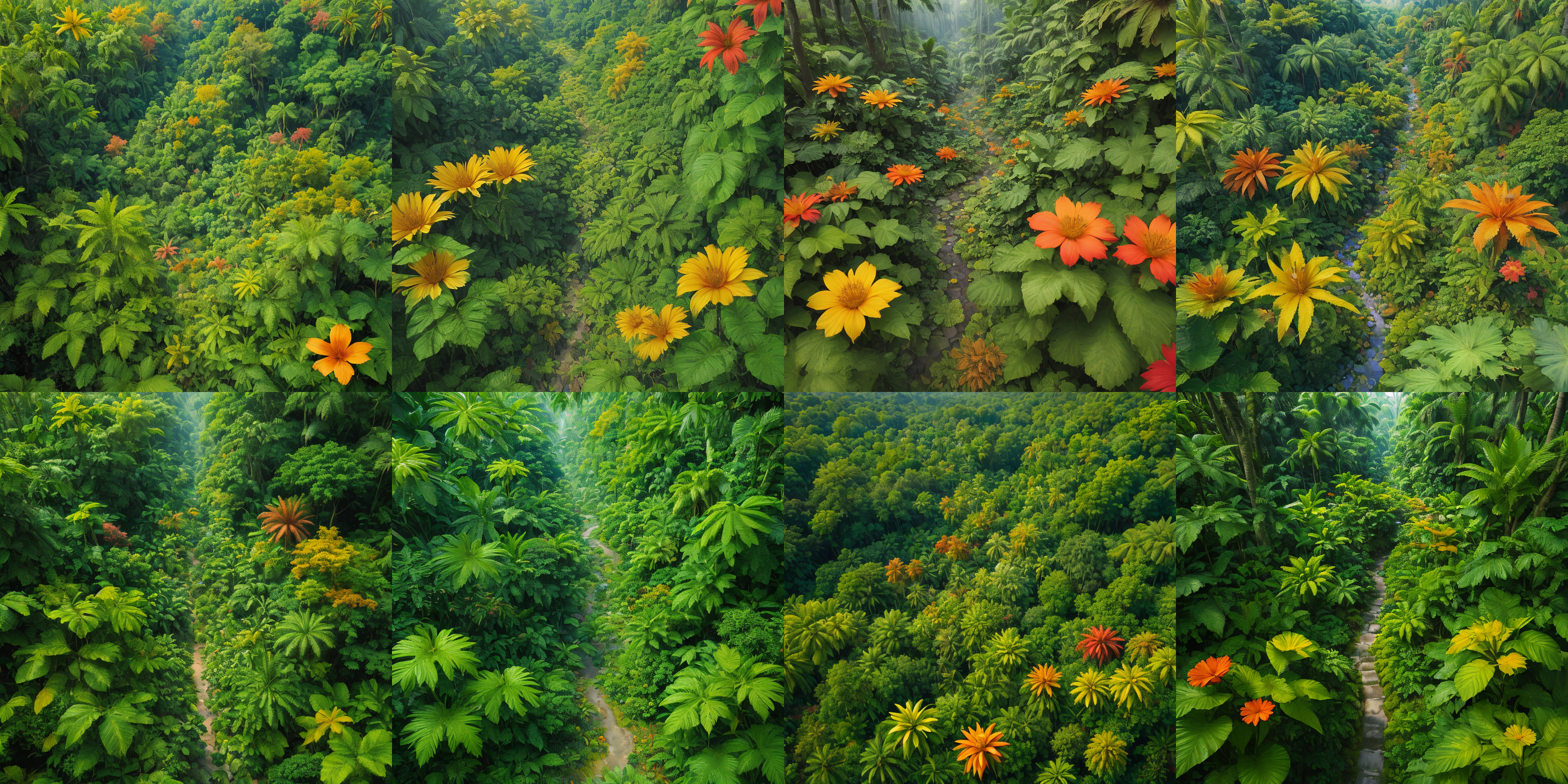}
        \caption{Aerial view of a vibrant tropical rainforest, filled with lively green vegetation and colorful flowers, sunlight piercing through the canopy, high contrast, vivid colors}
    \end{subfigure}
    
    \begin{subfigure}[b]{0.7\linewidth}
        \includegraphics[width=\linewidth]{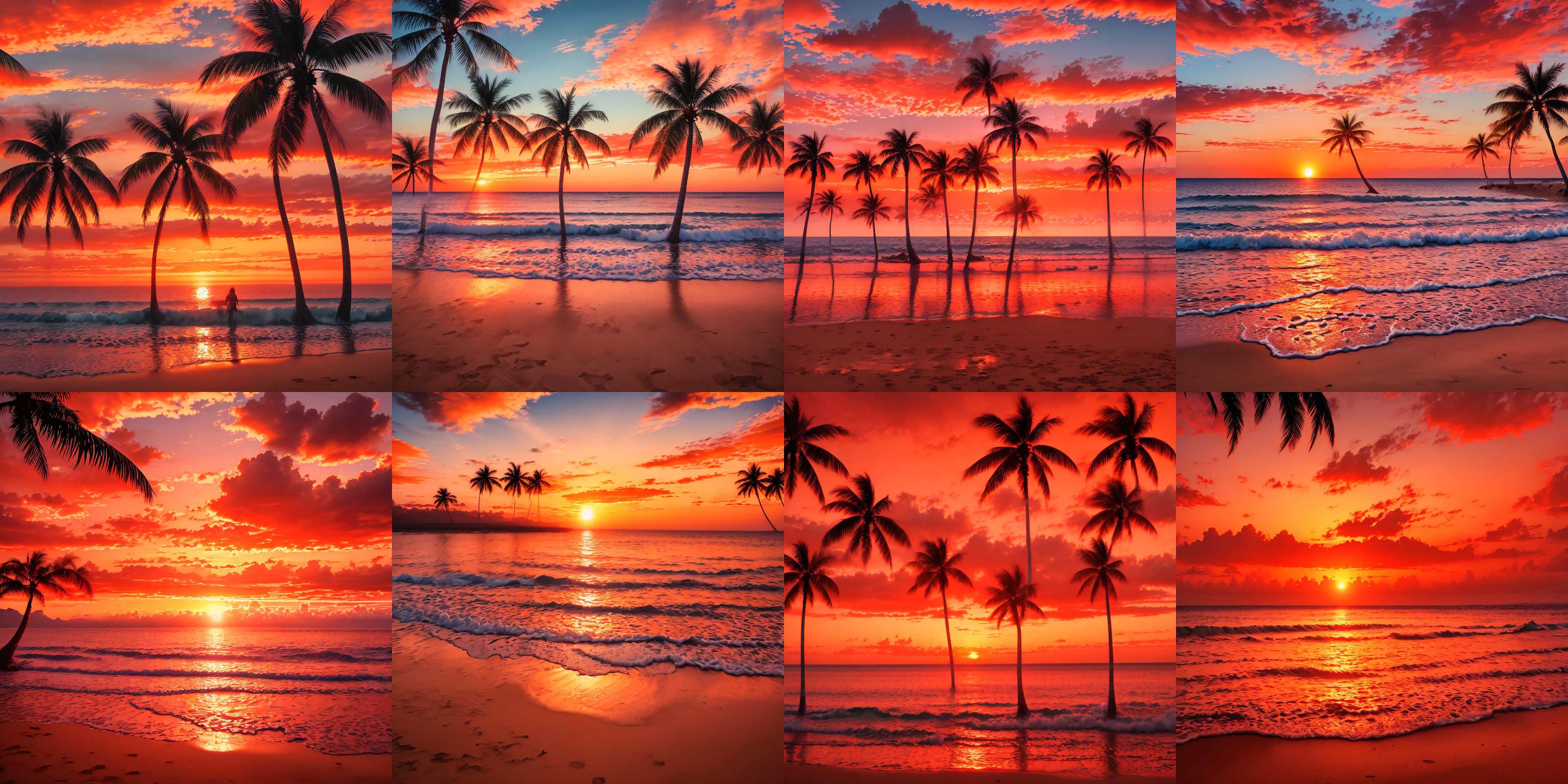}
        \caption{Tropical beach at sunset, the sky in splendid shades of orange and red, the sea reflecting the sun's afterglow, clear silhouettes of palm trees on the beach, high contrast, vivid colors}
    \end{subfigure}
    
    \begin{subfigure}[b]{0.7\linewidth}
        \includegraphics[width=\linewidth]{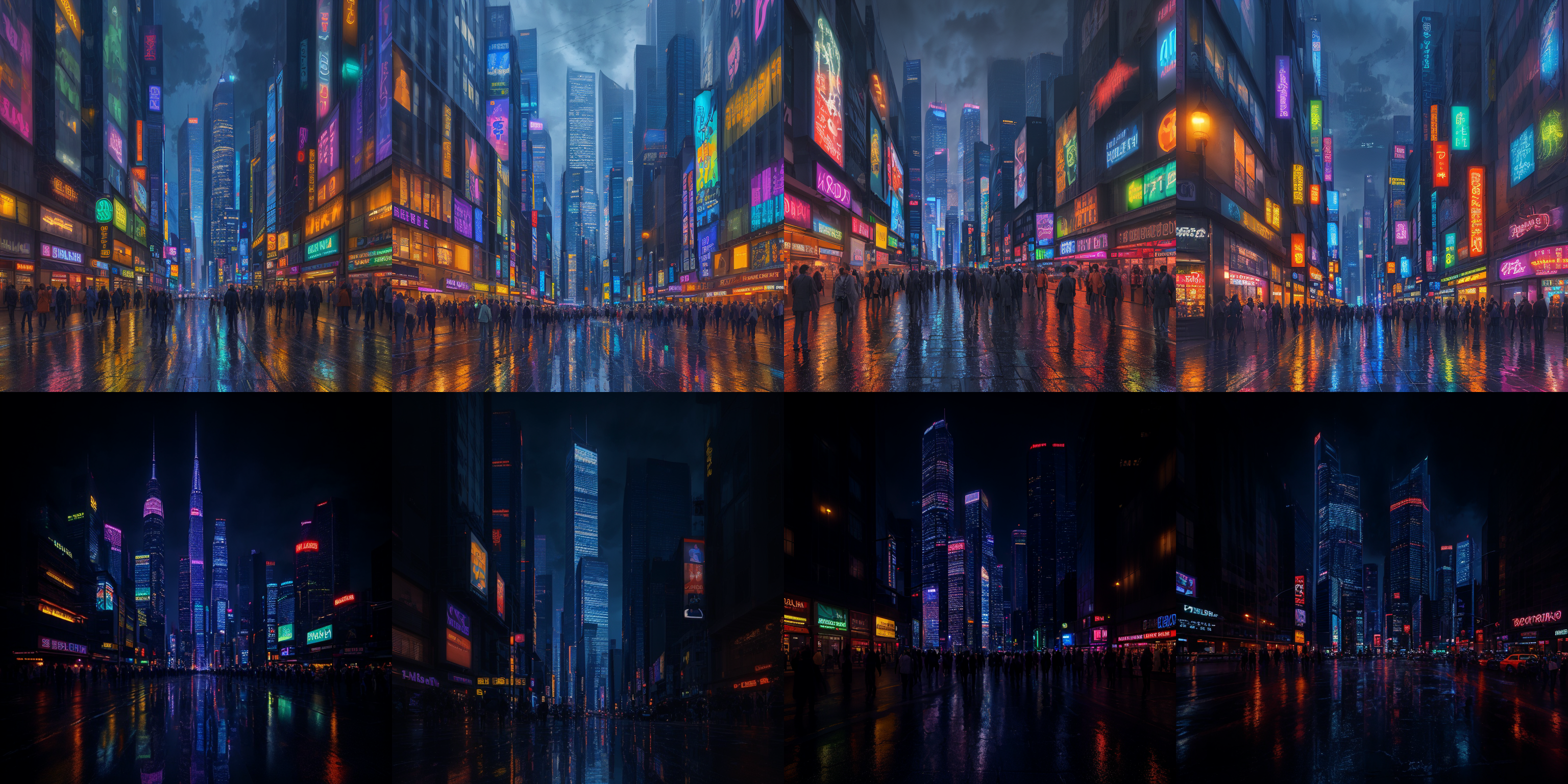}
        \caption{A cityscape at night with neon lights reflecting off wet streets, towering skyscrapers illuminated in a kaleidoscope of colors, high contrast between the bright lights and dark shadows}
    \end{subfigure}

    \caption{Additional Samples from LCM, top row from original model and bottom row with OMS.}
    \label{fig:lcm}
\end{figure*}

Here we demonstrate more examples based on SD1.5~\cref{fig:sd15}, SD2.1~\cref{fig:sd21} and LCM~\cref{fig:lcm} with OMS. In each subfigure, top row are the images directly sampled from raw pre-trained model, while bottom row are the results with OMS. In this experiment, all three pre-trained base model \emph{share the same OMS module}. 

\section*{Limitations}

We believe that the OMS module can be integrated into the student model through distillation, thereby reducing the cost of the additional step. Similarly, in the process of training from scratch or fine-tuning, we can also incorporate the OMS module into the backbone model, only needing to assign a pseudo-t condition to the OMS.  However, doing so would lead to changes in the pre-trained model parameters, and thus is not included in the scope of discussion of this work. 

\end{document}